\documentclass[sigconf,natbib=false,review]{sigkddExp}

\AtBeginDocument{%
  }
\usepackage{biblatex}
\usepackage{amsmath,amsfonts}
\usepackage{algorithmic}
\usepackage{graphicx}
\usepackage{textcomp}
\usepackage{xcolor}

\usepackage{algorithm}
\usepackage{subfigure}

\usepackage{diagbox}
\usepackage{threeparttable}
\usepackage{multirow}
\usepackage{color}

\usepackage{bbold}

\usepackage{graphicx}

\usepackage{makecell}

\usepackage{caption}

\usepackage{boldline}
\usepackage{colortbl}
\usepackage{wrapfig}
\usepackage{adjustbox}
\usepackage{enumitem}
\usepackage{booktabs} 
\usepackage{authblk}
\usepackage{setspace}
\usepackage[scaled]{helvet} % 使用 Helvetica 模拟 Arial 字体
\begin{document}

\newcommand{\wenbin}[1]{{\color{blue}{\small\bf\sf [Wenbin: #1]}}}
\newcommand{\avash}[1]{{\color{purple}{\small\bf\sf [Avash: #1]}}}

\title{Uncertain Boundaries: Multidisciplinary Approaches to Copyright Issues in Generative AI}
% \title{Fairness in Large Language Models: definitions, evaluation and algorithms}

% \author{Zhibo Chu}
%  \email{zb.chu@mail.ustc.edu.cn}
% \affiliation{%
%    \institution{University of Science and Technology of China}
%    \city{Heifei, Anhui}
%    \country{China}
% }

% \author{Zichong Wang}
%  \email{ziwang@fiu.edu}
% \affiliation{%
%    \institution{Florida International University}
%    \city{Miami, FL}
%    \country{US}
% }

% \author{Wenbin Zhang}
%  \email{wenbin.zhang@fiu.edu}
% \affiliation{%
%    \institution{Florida International University}
%    \city{Miami, FL}
%    \country{US}
% }
\balancecolumns
\numberofauthors{5}

% \author{
% %
% % The command \alignauthor (no curly braces needed) should
% % precede each author name, affiliation/snail-mail address and
% % e-mail address. Additionally, tag each line of
% % affiliation/address with \affaddr, and tag the
% %% e-mail address with \email.
% \alignauthor Zichong Wang\\
%        \affaddr{Florida International University}\\
%        \affaddr{Miami, FL, USA}\\
% \alignauthor Zhipeng Yin\\
%        \affaddr{Florida International University}\\
%        \affaddr{Miami, FL, USA}\\
% \alignauthor Fang Liu\\
%        \affaddr{University of Notre Dame}\\
%        \affaddr{Notre Dame, IN, USA}\\
%        \and
% \alignauthor Zhen Liu\\
%        \affaddr{Guangdong University of Foreign Studies}\\
%        \affaddr{Guangdong, China}\\
% \alignauthor Jun Liu\\
%        \affaddr{Carnegie Mellon University}\\
%        \affaddr{Pittsburgh, PA, USA}\\
% \alignauthor Xinda Wang\\
%        \affaddr{University of Texas at Dallas}\\
%        \affaddr{Dallas, TEX, USA}\\
%        \and
% \alignauthor Christine Lisetti\\
%        \affaddr{Florida International University}\\
%        \affaddr{Miami, FL, USA}\\
% \alignauthor Shuigeng Zhou\\
%        \affaddr{Fudan University}\\
%        \affaddr{Shanghai, China}\\
% \alignauthor Wenbin Zhang\titlenote{Corresponding author}\\
%        \affaddr{Florida International University}\\
%        \affaddr{Miami, FL, USA}\\
% }

\renewcommand\Authfont{\fontfamily{phv} \selectfont\fontsize{10}{12}} 
\renewcommand\Affilfont{\fontfamily{phv} \selectfont\fontsize{9}{11}}  
\fontsize{10}{12}
\author[1]{Archer Amon}
\author[1]{Zhipeng Yin}
\author[1]{Zichong Wang}
\author[1]{Avash Palikhe}
\author[1*]{\fontsize{9}{10}\vspace{-0.3cm} Wenbin Zhang}

\affil[1]{Florida International University, Miami, FL, USA}
\affil[ ]{\{aamon002, zyin007, ziwang, apali007, wenbin.zhang\}@fiu.edu}

\maketitle

\let\oldthefootnote\thefootnote            % 保存当前的编号方式
\renewcommand{\thefootnote}{\fnsymbol{footnote}}  % 临时改成符号样式
\footnotetext[1]{Corresponding author}   % 显示 *
\let\thefootnote\oldthefootnote            % 恢复编号

% \renewcommand{\shortauthors}{Trovato et al..}

\begin{abstract}

Generative AI is becoming increasingly prevalent in creative fields, sparking urgent debates over how current copyright laws can keep pace with technological innovation. Recent controversies of AI models generating near-replicas of copyrighted material highlight the need to adapt current legal frameworks and develop technical methods to mitigate copyright infringement risks. This task requires understanding the intersection between computational concepts such as large-scale data scraping and probabilistic content generation, legal definitions of originality and fair use, and economic impacts on intellectual property (IP) rights holders. However, most existing research on copyright in AI takes a purely computer science or law-based approach, leaving a gap in coordinating these approaches that only multidisciplinary efforts can effectively address. To bridge this gap, our survey adopts a comprehensive approach synthesizing insights from law, policy, economics, and computer science. It begins by discussing the foundational goals and considerations that should be applied to copyright in generative AI, followed by methods for detecting and assessing potential violations in AI system outputs. Next, it explores various regulatory options influenced by legal, policy, and economic frameworks to manage and mitigate copyright concerns associated with generative AI and reconcile the interests of IP rights holders with that of generative AI producers. The discussion then introduces techniques to safeguard individual creative works from unauthorized replication, such as watermarking and cryptographic protections. Finally, it describes advanced training strategies designed to prevent AI models from reproducing protected content. In doing so, we highlight key opportunities for action and offer actionable strategies that creators, developers, and policymakers can use in navigating the evolving copyright landscape.

\end{abstract}

\keywords{Generative AI, Copyright law, Intellectual property, AI policy, Copyright infringement detection}

% \renewcommand{\thefootnote}{}
% \footnotetext{* Corresponding author
% }
% \footnotetext{Email: \{ziwang, wenbin.zhang\}@fiu.edu}
% \let\thefootnote\relax\footnotetext{* Corresponding author}


\section{Introduction}
\label{sec:introduction}
\vspace{0.4cm}

% \wenbin{add citations -- Done}
% \wenbin{fix single/double quotes syntax issues. -- Done}

% The growing popularity of generative AI has reignited concerns around intellectual property, particularly due to how many commercial AI models are trained on datasets scraped freely from the web. This practice has led to several high-profile cases where these models produced work nearly identical to copyrighted content~\cite{Stempel_2023,visPlagiarism,thompson_2024}. In response, some companies have attempted to deflect responsibility from developers by blaming users who prompt models to generate infringing content~\cite{midjourneyTerms}. Others have relied on broad interpretations of `fair use' to dismiss claims of copyright infringement, further complicating the chain of accountability. This ambiguity not only undermines the rights of copyright holders but also diminishes the quality and credibility of generative AI systems. Although technical methods to prevent these violations exist, their adoption has been slow. Technical efforts are further complicated by the rapidly-changing landscape of AI development, and a lack of clear standards for assessing whether common industry practices do or do not violate copyright law.  There is also a notable lack of resources that comprehensively review these solutions across all stages of AI development, making it difficult for the many stakeholders involved in designing, creating, deploying, and regulating these generative AI systems to coordinate their efforts.  
The growing popularity of generative AI has reignited significant concerns around intellectual property (IP) rights, especially given that many commercial AI models rely heavily on datasets freely scraped from the internet~\cite{chesterman2024good}. This practice has resulted in several high-profile controversies, including real-world cases such as the legal dispute involving Stability AI's Stable Diffusion model allegedly replicating artists styles without permission~\cite{hayes2023generative}, and lawsuits filed by artists against companies like Midjourney and DeviantArt for unauthorized use of their works~\cite{frosio2024generative}. Additionally, literary authors have raised objections against large language models like ChatGPT for generating content closely resembling their copyrighted texts~\cite{Stempel_2023,visPlagiarism,thompson_2024}. Companies have responded to these incidents in various ways: some have attempted to shift blame onto users who prompt models to create potentially infringing content~\cite{midjourneyTerms}, while others invoke broad and ambiguous interpretations of ``fair use'' to defend their practices, further complicating accountability and enforcement~\cite{cotter2007fair}. This ambiguity not only weakens the legal protections available to copyright holders but also undermines public trust and the perceived integrity of generative AI systems. Although technical measures to proactively mitigate these violations exist, such as content fingerprinting, watermarking, and cryptographic methods, their adoption has been slow and inconsistent~\cite{zhang2024research}. Additionally, the rapidly evolving AI technology landscape and the absence of clear legal standards exacerbate these challenges, making it difficult for stakeholders, including creators, developers, and regulators, to effectively coordinate their efforts.

%\wenbin{four goals including legal guidance/support? --- Done}

Addressing these challenges requires a holistic approach that combines technical solutions with supportive policy frameworks. To navigate these complexities, this paper aims to provide a comprehensive survey of current methods for enhancing copyright compliance in generative AI, focusing on four main goals: (1) detecting copyright violations and evaluating model performance, (2) understanding how regulatory landscapes shape technical strategies for protecting individual copyrighted works from unauthorized use, (3) protecting individual copyrighted works from being used in AI systems without authorization, and (4) designing AI models in a way that prevents generation of content violating copyright. In doing so, we adopt a uniquely multidisciplinary perspective, incorporating insights from computer science, law, policy, and economics to provide a more holistic framework for addressing copyright challenges in generative AI. Within this framework, we evaluate various methods based on their effectiveness and feasibility in preventing copyright violations, while also considering their impact on the utility of generative AI models. Through this survey, we aim to provide actionable insights for the development of technical, legal, and policy strategies, enabling creators, developers, and policymakers to navigate the complex copyright challenges introduced by generative AI.
% To ensure a comprehensive understanding of how these methods function in practice, we organize our analysis around key stages of the AI supply chain~\cite{lee2024supplychain}, including systematically tracing copyright considerations from the initial creation of protected works through dataset compilation, model training, and the deployment of final AI systems.
% Alongside these goals, we also evaluate the effectiveness and feasibility of various methods aimed at preventing copyright violations while preserving the usefulness of generative AI models. To ensure a comprehensive understanding of how these methods operate in practice, we frame our analysis within the stages of the AI supply chain~\cite{lee2024supplychain}, systematically tracing copyright considerations from the initial creation of protected works through dataset compilation, model training, and finally, the deployment of finished AI systems.

\noindent \textbf{Paper Structure.}
The subsequent sections of this paper are structured as follows: Section 2 provides foundational background, explaining key concepts of generative AI and the complexities of copyright law as it applies to AI-generated content. Section 3 outlines a comprehensive taxonomy categorizing multidisciplinary methods addressing copyright issues in generative AI. Section 4 presents techniques for detecting and evaluating potential copyright infringements by AI models. Section 5 reviews regulatory frameworks and policy measures designed to manage and mitigate copyright risks associated with generative AI. Sections 6 and 7 discuss technical approaches aimed at protecting copyrighted content from unauthorized AI copying, as well as advanced training methods specifically designed to prevent generative AI models from producing infringing output. Finally, Sections 8 and 9 explore available resources, ongoing challenges, and future research directions in addressing copyright concerns within the evolving landscape of generative AI.

\section{Background}
\label{sec:Background}
\vspace{0.3cm}

\subsection{Generative AI Models}
\vspace{0.3cm}

\subsubsection{Definition}
\vspace{0.4cm}
Generative AI broadly refers to artificial intelligence systems capable of creating new content, such as text, images, audio, or video, by learning patterns from extensive datasets. Compared to traditional AI systems, which focus on analyzing existing data or making decisions, generative AI models produce new outputs in response to user prompts~\cite{gozalobrizuela2023}. These models are typically trained on large datasets, enabling them to generate content that mimics the style and structure of the data they were trained on, although small models are also an emerging field of research~\cite{xu2023small,hsieh2023small,li2024nsmall}. Common model types include text-to-text, text-to-image, text-to-video, and image-to-video AI, as well as multimodal systems~\cite{kang2023multimodal,liang2024multimodal,lu2024multimodal} that integrate multiple input and output forms.

\subsubsection{Model architecture}
\vspace{0.4cm}
These models are typically trained on massive amounts of data, and utilize large architectures to encode inputs into a high-dimensional latent space and use a generator model to produce varied outputs through a stochastic behavior~\cite{gozalobrizuela2023}. Most generative AI systems today are built on a transformer architecture consisting of an encoder and decoder, using a multi-head self-attention mechanism to handle long-term dependencies in data by assigning higher weights to more relevant tokens~\cite{cao2023survey}. Because of the large amount of resources needed, generative models are often trained through a ``pre-training'' paradigm, where general purpose models are later fine-tuned for specific applications, creating a more complex chain of command where issues of indirect liability are more likely to come into play~\cite{yew2024liability}.  

\subsection{Copyright Applied to Generative AI}
\vspace{0.3cm}

\subsubsection{Fair use standards}
\vspace{0.4cm}
Broadly speaking, copyright allows the creator of an original work to prevent others from creating and profiting from unauthorized replication, distribution, or derivation of their works~\cite{defintion-copyright-office}. When evaluating copyright violations, US law has identified four main pillars that constitute ``fair use'' of copyrighted material: (1) the purpose and character of the use, (2) the nature of the copyrighted work, (3) the amount and substantiality of the portion taken, and (4) the effect of the use on the potential market for the work~\cite{fairUseDef}. These questions are particularly difficult to address in the context of generative AI systems as copyrighted data is often compiled into massive datasets for model training, and the impact of a particular copyrighted work on generated outputs cannot always be traced clearly. 

\subsubsection{Analyzing violations}
\vspace{0.4cm}
Given the complexity of understanding fair use, analyzing potential copyright violations in generative AI often requires mixing legal and technical understandings. While there is some legal precedent for copyright cases involving search engines, web code, and APIs~\cite{Opderbeck2024}, many questions have yet to be answered in the context of AI systems. There is no clearly delineated amount for what counts as ``fair use'' of a copyrighted work; for example, the copyright of certain content types such as cookbooks or dictionaries is more likely to be infringed by copying even a small part, whereas this may be acceptable for larger novels~\cite{henderson2023Foundation}. AI also frequently combines both expressive and non-expressive properties~\cite{Pasquale2024}, and analyzing compliance often requires looking at both low-level content transformations such as n-grams and verbatim copying, and higher level concepts like themes and storylines~\cite{henderson2023Foundation}.

\subsubsection{Levels of memorization} % possibly change name "Why relate to copyright issues" / "Why copyright emerges"
\vspace{0.4cm}
When searching for technical evidence to explain AI copyright violations, many researchers point to the phenomenon of memorization, where generative AI models reproduce near-exact copies of training data~\cite{cooper2024mem}. Memorization can arise due to overfitting or the underlying distribution of data~\cite{ross2024mem}, and leads to both direct verbatim reproduction and more subtle forms of copying. Yet despite language models frequently committing plagiarism at the paraphrase or idea-based level~\cite{langModelPlag}, most existing research focuses only on verbatim copying, making more subtle forms harder to assess ~\cite{copybench}.  There is thus a need to expand AI copyright research into identifying and mitigating indirect copying of protected works.

\subsubsection{Practical challenges for the AI context}
\vspace{0.4cm}
Another factor that often complicates AI copyright issues from both a legal and technical perspective is the opaque and decentralized nature of AI development. Many generative AI models are ``black-box'', meaning developers and researchers cannot fully understand their internal functions and trace how copyrighted content may be appearing~\cite{li2021blackbox,Schneider2024}. AI development today also occurs through a highly distributed supply chain~\cite{lee2024supplychain}, and lack of coordination between involved parties reduces the chance for methods to be effective at scale or resilient across later transformations of an AI model~\cite{Ohm2024}.

\nocite{wang2023preventing, zhang2023individual,wang2023fg2an,yazdani2024comprehensive,wang2023mitigating,chinta2023optimization,wang2024history,chu2024fairness, dzuong2024uncertain,yin2024improving,wang2023towards,wang2024toward,chinta2024fairaied,doan2024fairness,wang2024individual,doan2024fairness1,wang2024advancing,wang2024group,wang2024individual,yin2024accessible,wang2025fg,wang2025graph,wang2025fair,wang2025towards,yin2025digital,chinta2025ai,wang2025fdgen,wang2025fairgnn,wang2025limiteddemographics,wang2025Fairness,wang2025Redefining,zhang2019faht,zhang2024ai,zhang2022longitudinal,zhang2023censored,zhang2025fairness,zhang2022fairness}

\section{Taxonomy Outline}
\label{sec:Taxonomy Outline}
\vspace{0.4cm}

Facing dual challenges from legal and technical environments, many researchers have called for a co-evolution of technology and law so that developments in each field may support the other~\cite{henderson2023Foundation}. Considering this need for interdisciplinary work, our paper aims to combine a detailed overview of state-of-the-art technical methods for reducing copyright violations in generative AI with an analysis of the regulatory landscape which may support these methods. Specifically, we organize our analysis using a framework that maps copyright concerns across key stages of the generative AI lifecycle, integrating legal and technical work into a coherent structure. As shown in Figure 1, we categorize recent work supporting copyright compliance for generative AI into four specific focus areas, tracing copyright issues from foundational challenges to practical solutions while incorporating a diverse range of actors and development stages. We begin with \textbf{(1) detecting and assessing copyright violations}, outlining methods that identify where and how infringement occurs so that targeted solutions can be applied. Next, we discuss \textbf{(2) regulatory approaches} that can help facilitate more effective copyright protection in AI. In the following two sections, we explore technical mitigation strategies in two distinct areas: Methods for \textbf{(3) protecting copyrighted works}, which give individual creators tools to safeguard their from unauthorized use, and \textbf{(4) preventing copyright infringement}, covering model-level training strategies aimed at reducing the over likelihood of AI systems generating infringing outputs. After covering these four areas, we present resources like datasets and toolkits to support these goals, and conclude by highlighting emerging challenges in the field.

\begin{figure}
    \centering
    \includegraphics[width=1\linewidth]{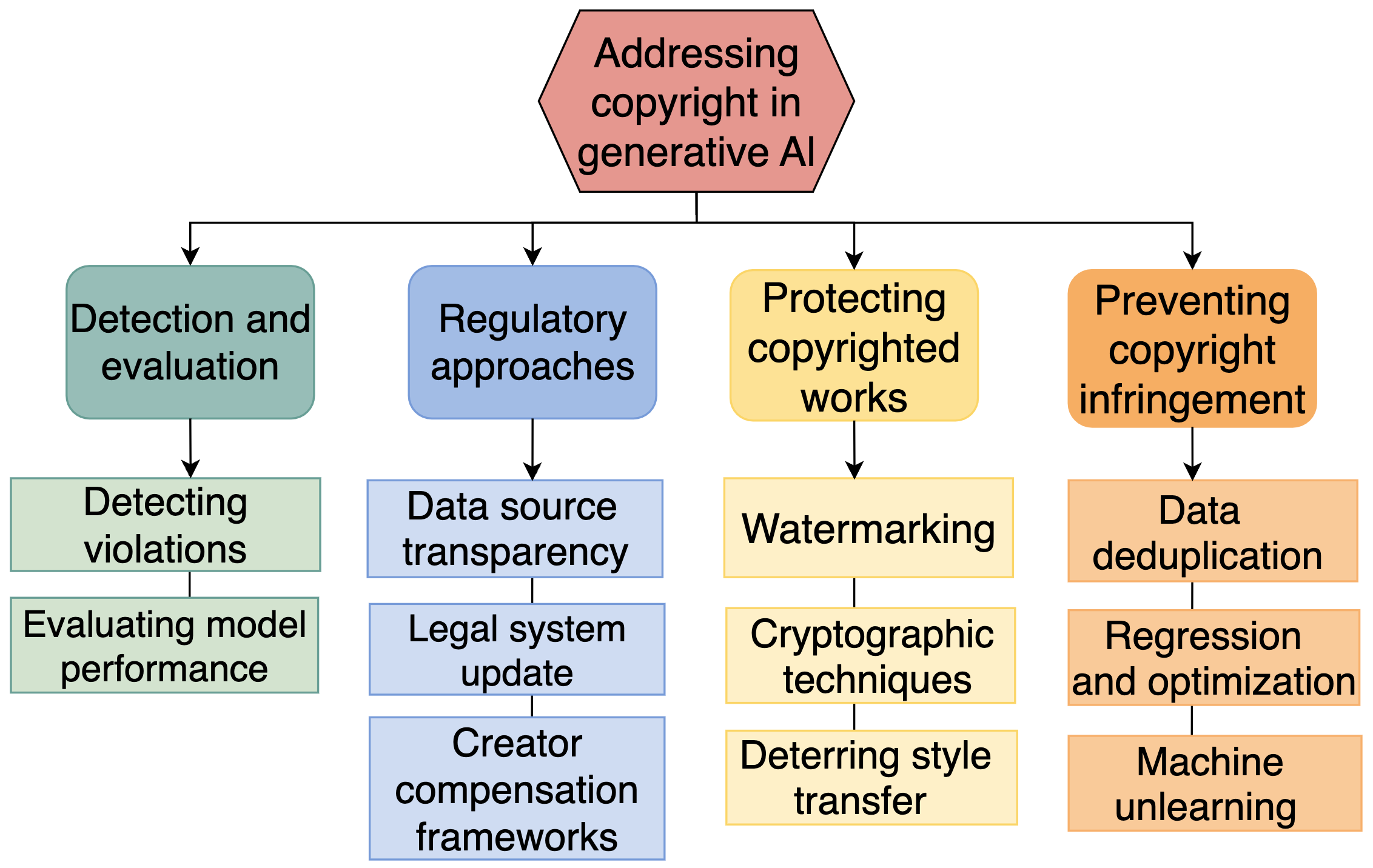}
    \caption{An overview of our proposed taxonomy.}
    \label{fig:Taxonomy}
\end{figure}

%\wenbin{figure particularly the font too small. --Done}

\section{Detecting and Evaluating Violations}
\label{sec:Detecting and Evaluating Violations}
\vspace{0.4cm}
We begin by surveying methods for detecting and evaluating instances where AI models are likely to infringe on copyright. These methods are critical as they help stakeholders to identify problem sources and apply targeted remedies, assess risk and compliance, and use concrete evidence of infringement to inform legal and policy decisions. We divide the methods here into two main categories: (1) web tools for detecting unauthorized AI-enabled reproduction of copyrighted works and (2) datasets and methods for evaluating general model performance.

\subsection{Detecting violations}
\label{sec:Detecting violations}
\vspace{0.4cm}

Methods for flagging AI-enabled copyright violations draw from two primary areas of research: detecting copyright infringement and detecting AI-generated content. There is often a trade-off between the two, as methods for detecting infringement often have limited applicability once content has been altered from its original form, and AI-generated content (AIGC) detection relies on picking up special signatures from AI-generated content which may be less present in a close copy of a human creator's work. 

% \wenbin{fix illegal empty space throughout the paper by reading it through: ''Google Reverse Image Search'' -- Zhipeng :Done}

\subsubsection{Identifying copyright infringement}
\vspace{0.4cm}
The first set of methods focuses on finding places where protected content is replicated without authorization, primarily relying on simple web tools. Reverse image search tools such as Yandex\footnote{\url{https://yandex.com/images}}, Tineye\footnote{\url{https://tineye.com}}, and Google Reverse Image Search\footnote{\url{https://google.com/images}}, along with text search tools like Scribbr\footnote{\url{https://scribbr.com/plagiarism-checker/}} and Grammarly\footnote{\url{https://grammarly.com}} allow users to find instances where content may have been replicated. Recent models have also emerged which leverage AI to search the web for potential violations, combining BERT with DNN models to search for and flag infringing content~\cite{HernandezSuarez2023}. However, many research studies on detecting copyright infringement are highly limited in scope, and may fail at detecting violations made by AI which alters the content beyond its original form~\cite{langModelPlag,copybench}. Special applications where the expressive and non-expressive elements of a work are closely linked, like with LLM-powered code generation, may also need unique methods for determining if a certain use counts as infringement or not~\cite{rafatijo2020software,xu2024license}.

\subsubsection{Identifying AI Generated Content}
\vspace{0.4cm}
The next set of methods aims to detect AI generated content and trace it back to its original source to identify potential violations. A wide variety of methods have been developed to detect AI generated content across text~\cite{xie2024mugc,Elkhatat2023}, image~\cite{ImageAIGC}, video~\cite{he2024aigcvid}, and multi-modal~\cite{huang2024ruai,yu2024aigc} content.  They show that traditional methods such as logistic regression, random forest, SVM, and classifier-based methods display a fair level of accuracy at separating human from AI created content. However, not every AI generated image is one that infringes on copyright, so it is also necessary to have methods for scanning content registries to identify what images it was likely sourced from.

\subsubsection{Fingerprinting for Comparison}
\vspace{0.4cm}
Fingerprinting has been recommended for many applications to better enable identification and take-down of copyrighted material, providing a unique trace allowing digital content to be identified and attributed to its source~\cite{decentralizedFingerprinting}. Preetha and Bindu use a wavelet based video fingerprint to extract signatures from different images created from a video, extracting both temporal and spatial features into a compact form which can be stored in a video database and used to determine whether a query video is drawn from that database source~\cite{Preetha2023}. Ning et al. develop a similar system allowing users to register content using its fingerprint rather than the original work~\cite{ning2021reg}. While many different strategies for using digital fingerprints have been developed, they generally share four main characteristics: uniqueness, stability, extractability, and compactness~\cite{chen2024fingerprint}. As a result, fingerprints may provide an efficient way to identify infringing works at scale, but current research applied specifically to the context of generative AI problems is limited. 

\subsection{Assessing model performance}
\vspace{0.3cm}
\subsubsection{Jailbreaking methods} 
\vspace{0.4cm}
Several methods aim to test how easily a model can be made to generate protected content. Text-to-image models can often be prompted to generate copyrighted content even when keywords for a protected IP are replaced with a description of the image~\cite{kim2024jail}. Kim et al. introduce an automatic prompt generation pipeline using LLMs to autogenerate descriptions and create revised prompts designed to induce reproduction of copyrighted content~\cite{kim2024jail}. Their model can evaluate LLM copyright compliance without requiring access to any internal weights, allowing it to function on black-box systems. The prompts generated through their pipeline successfully jailbreak ChatGPT to generate copyrighted content 76\% of the time, with an 11\% block rate. Test cases can be designed for exploring a model's tendency to engage in specific copying behaviors, like verbatim vs indirect copying.

\subsubsection{AI dataset probing}
\vspace{0.4cm}
Similar methods aim to reduce the black-box nature of AI models by probing into whether certain copyrighted content is included in their training dataset. For this purpose, Duarte et al. introduce DE-COP, a benchmarking method designed to determine if a piece of copyrighted content is included in a training dataset~\cite{duarte2024decop}. Their approach is to probe an LLM with multiple-choice questions to complete a passage of a suspected target book, with options including both verbatim text and paraphrases. The LLMs that were tested showed substantially different performance on these tests depending on whether or not a book is in their training dataset, with the DE-COP method showing a 72\% accuracy for detecting suspect books, compared to about 4\% for previous models. However, information about whether or not something is in a model's dataset is not enough to prove a violation of copyright. This method should thus be combined with other strategies which examine model outputs to find violations. 

\subsubsection{Model-level risk quantification} 
\vspace{0.4cm}
To fulfill the need for a broad metric for evaluating copyright risk, Zhou et al. introduce CopyScope, a framework for quantifying infringement at the model level by using Fréchet Inception Distance (FID) to capture image similarity in the way that most closely mirrors human perception~\cite{zhou2023copyscope}. Using an ensemble based approach of trying different combinations of model sub-components, Zhou et al. use FID-Shapley values to calculate how much each component contributes to the final image's likeness to the training data, identifying which models are most likely to cause infringement issues and should be identified as targets for extra attention.

\section{Regulatory Options for AI Copyright Issues}
\vspace{0.4cm}

\sloppy
Copyright challenges in generative AI cannot be solved through advancements in computing alone; disciplines such as law and policy play a central role in shaping discussions around copyright and promoting specific values in design. Legal and policy standards not only work to clarify where violations are present~\cite{Rosati2024}, but also frequently serve as a prerequisite for implementing technical safeguards on a wide scale. For instance, several mitigation techniques depend on a well-documented data life cycle, requiring clear transparency about where data is sourced and how it is used~\cite{zhang2024lifecycle}, which better regulations can help coordinate. These frameworks also work to shape broader industry practices around data collection, creator consent, and compensation, fundamentally affecting what copyright issues emerge. Taking a uniquely interdisciplinary lens, this section explores key regulatory approaches that interact with AI copyright mitigation and enforcement, integrating perspectives from law, economics, and more.

%\wenbin{fix citation issue above-- read the paper through to fix such issues. -- Done}

\subsection{Data source transparency}
\vspace{0.3cm}
\subsubsection{Voluntary measures and Industry standards}
\vspace{0.4cm}
Transparency in AI training data is a foundational principle for advancing copyright protection in AI~\cite{RodriguezMaffioli2023}. Transparency measures empower copyright holders to monitor and protect against misuse of their works, and promote better principles and processes for designing AI models. Inspired by similar ``privacy-by-design'' principles, several experts advocate for adopting measures that actively promote transparency throughout the AI development process~\cite{Felzmann2020}. Groups like the Coalition for Content Provenance and Authenticity and the Content Authenticity Initiative have frequently contributed to these discussions, developing technical standards for tracking data origins with features that allow rights holders to specify whether training is allowed~\cite{RodriguezMaffioli2023}. However, the adoption of such measures remains largely dependent on the voluntary actions of individual companies, which limits their widespread effectiveness.

\subsubsection{Standardized labels}
\vspace{0.4cm}
Another approach calls for standardized labels to harmonize transparency efforts. These AI ``nutrition labels'', inspired by similar efforts in the food agency, call for disclosing information on an AI system’s data sources, potential risks, and limitations~\cite{chmielinski2022nutrition}. While these models have been adapted to support labeling of generative AI~\cite{si2024nutrition}, more work is necessary to explore how they can be applied to the specific context of copyright. A similar proposed tool is ``data cards'', or structured summaries that provide essential facts about datasets and explain the rationale behind decisions made in creating them~\cite{Pushkarna2022}. Professional organizations and regulatory authorities could adopt policies to promote the adoption of these measures or integrate them within other documentation and reporting requirements.

\subsubsection{AI system audits}
\vspace{0.4cm}
Independent audits of AI systems may be proposed as part of a broader licensing framework or an industry standard certification separate from statutory policy. Auditing for copyright is typically done on model data sets to understand the data collection process and nature of the dataset~\cite{manheim2024audit}. However, these ``data audits'' typically focus on general industry data practices rather than holding dataset creators accountable, and are often divorced from other model-level audits, creating a disconnect between data understanding and deployment regulations~\cite{birhane2024audit}. Noting the fragmentation of the current AI audit landscape, Manheim et al. recommend creating AI audit standards boards keep auditing standards up to date with current advancements in AI and clarify which standards are suited for specific domains and applications~\cite{manheim2024audit}. This work also has a technical dimension, as decisions need to be made about how much access should be given to auditors. In this scope, Casper et al. propose expanding beyond traditional ``black-box'' access audits which only observe AI system outputs. To make audits more robust, they suggest including ``white-box'' access to information about model weights and inner workings, and ``outside-the-box'' access that provides information about training and deployment processes~\cite{Casper2024}.

\subsection{Copyright legal system updates}
\vspace{0.4cm}
%\subsubsection{Considerations for legal change}

Considerations for legal change. As generative AI systems continue to increase their capabilities and become more widespread, existing doctrines of copyright law become increasingly unsuited for resolving disputes about AI~\cite{alhadeff2024fairuse,kivus2024law}. This creates a need for broader structural change, which is likely to be implemented through a gradual system of legislative action and interpreted rulings rather than a single policy. Many ideas revolve around strengthening protections for creators~\cite{Geiger2024} or updating the copyright law system to address specific questions around data scraping and ownership in the digital age ~\cite{chesterman2024good}.  In these discussions, policymakers often need to manage the trade-off between transparency and feasibility of implementation. Beyond facing criticisms of stifling innovation which may turn government support against these measures~\cite{chan2024innov}, creating too harsh of standards around training data may significantly limit the amount available, and limited data is more likely to reinforce biases and stereotypes~\cite{Geiger2024}. 

\subsubsection{Opt-out policies}
\vspace{0.4cm}
Opt-out provisions aim to reconcile issues of consent and copyright by allowing creators to opt out of having their work used for AI training~\cite{zhang2024opt}. However, it is often unclear how they are meant to function in practice~\cite{ziaja2024opt}. Existing opt-out methods are often difficult and seen by creators as largely a PR stunt~\cite{Pasquale2024}, causing a need for more research to address the fragmentation of opt-out policies in the status quo~\cite{keller2023opt}. Pasquale and Sun (2024) propose a mandatory opt-out mechanism requiring AI developers to remove works from their databases upon request if infringement has been documented, and take action action to prevent the infringing content from being made again~\cite{Pasquale2024}. While this method creates more transparency by requiring developers to properly manage their datasets and confirm upon request if a source has been used~\cite{Pasquale2024}, it functions more as a post-hoc right to remove copyrighted content after violation has occurred, leaving a remaining need for more preventative opt-out strategies. Many rights holders may also be unaware of unauthorized AI reproductions of their work or lack information about opt-out procedures, creating barriers to establishing a system of full consent. 

\subsubsection{Content management registries}
\vspace{0.4cm}
To help streamline consent management for new AI training, Balan et al. introduce a decentralized registry for content creators to assert their right to opt in or out of AI training, combining distributed ledger technology with visual fingerprinting~\cite{balan2023opt}. Their prototype model prevents a scalable way to trace generative AI training data to determine consent, similar to many of the methods discussed in the protection section. They propose such a registry also be used to track information for compensating creators that opt in to AI training. 

\subsection{ Compensation frameworks}
\vspace{0.4cm}
Many new proposals have identified that reworking current compensation frameworks may help reconcile the interests of creators and AI producers. Compensation not only ensures individual creators are treated fairly, but also establishes better data provenance tracking and reduces power differentials between AI developers and creators~\cite{Pasquale2024} when frameworks are well designed. Compensation systems could be arranged through either new statutory~\cite{Geiger2024} or existing contract-based structures~\cite{wang2024econ}, working with collective rights management organizations to streamline negotiating and distributing remuneration. These models generally fall within three main categories, as illustrated in Table 1.

\begin{table}[h!]
\centering
\caption{Compensation frameworks for AI training data.}
\begin{tabular}{|>{\raggedright\arraybackslash}p{3.5cm}|>{\raggedright\arraybackslash}m{4.1cm}|} \hline 

\textbf{Compensation Model} & \textbf{Pays Based On} \\ \hline  
Pay to Train & Percentage of training data \\ \hline  
Pay to Train and Inspire& Contribution to generated outputs \\ \hline  
AI Royalties & Negotiated IP partner framework \\ \hline 
\end{tabular}

\end{table}

\subsubsection{Pay-to-train compensation}
\vspace{0.4cm}

The pay-to-train model involves rewarding IP holders based on the percentage of their contribution to a dataset. Such a model could use increased dataset transparency requirements to calculate payments based on the amount of copyrighted material within AI datasets and the monetary value attributed to its use for AI training. These payments could be afforded to individual creators, or distributed into collective funds to support creators. However, dataset sources are not always well-documented, and payments may be minimal for individual creators beyond famous authors or artists whose work is more likely to comprise a large portion~\cite{della_giustina_2024}. 

\subsubsection{Pay-to-train-and-inspire compensation}
\vspace{0.4cm}
This model works backwards to understand which items in a model's training data likely inspired a particular output, and distribute compensation accordingly~\cite{ducru2024royalties}. Wang et al. combine probabilistic methods with Shapley value interpretability techniques under a game theory model to establish a base framework for compensation in this way and show its viability through practical experimentation with common data sources~\cite{wang2024econ}. This method works best for AI models trained on limited data with copyright split between a smaller amount of owners. In some cases, it may be more practical to estimate aggregate payoffs for copyright owners across all generated outputs rather than tracing each output back to its source for compensation.

\subsubsection{AI royalties}
\vspace{0.4cm}
The last model, AI royalties, aims to create collaborative partnerships between IP rights holders and AI companies for compensation based on the market usage and value of their systems~\cite{ducru2024royalties}. This model could be implemented using existing contract law systems, recognizing the exclusive right of rights holders to commercial use of substantially similar copyright or trademark uses, and granting the AI company the right to use its IP to create outputs under defined use guidelines, sharing in a portion of the overall revenue generated by the system. Ducru et al. argue that this model is the best suited to mutually benefit the interests of AI creators and IP rights holders, and eliminates the need for case-by-case determinations by allowing a broader, predefined agreement that covers all outputs generated by the system~\cite{ducru2024royalties}.

\section{Protecting Copyrighted Works from AI}
\vspace{0.4cm}
We now introduce technical methods for mitigating copyright infringement in generative AI, starting with methods for protecting individual copyrighted works. These methods are primarily applied at the data collection and pre-processing level, and may be used by creators to protect their works as well as developers seeking to preserve attribution and traceability of training data used. While many protection techniques are frequently used in conjunction, we divide them into three primary categories: (1) watermarking, (2) cryptographic methods, and (3) deterring style transfer.

\subsection{Watermarking}
\vspace{0.4cm}
Watermarking methods serve to embed an imperceptible identification layer onto images or videos to identify ownership over the content. They can be embedded directly into inputs such as images, or their intermediate representations such as encoder/decoder feature maps~\cite{deng2024defense}. We divide this section into two main approaches based on the methods used to apply the watermark: attention-based mechanisms and cryptographic methods.

\subsubsection{Attention based mechanisms}
\vspace{0.4cm}
\sloppy
Attention-based techniques leverage AI models' tendency to focus on specific image regions to create watermarks which remain present after a high degree of manipulation. For example, Zhang et al. introduce deep learning-based video watermarks using a custom attention mechanism designed to recover a 32-bit watermark with 99\% accuracy post-transformation~\cite{robustVidWatermark}. Similar methods can be used for image content by applying watermark algorithms based on Tchebichef moments which describe the spatial distribution of an image's intensity~\cite{tchebichefWatermarking,imageTchebichef,Xiao2012tchebi}. By partitioning the host image into non-overlapping blocks and calculating Tchebichef moments for each block, Ernawan and Kabir are able to prioritize watermark embedding in areas of lower visual entropy or complexity. As a result, their method demonstrates resilience against noise disruptions and JPEG2000 compression, while preserving a high peak signal-to-noise ratio (PSNR) of approximately 40 dB. While attention mechanisms like these are more resilient to various transformations, testing may be needed to ensure watermarks maintain high enough `pattern uniformity' to be learned and reproduced by generative diffusion models~\cite{diffusionShield}.  

\subsubsection{Cryptography-based watermarks}
\vspace{0.4cm}
In order to further increase traceability and security, several techniques for watermarking incorporate various cryptographic methods. For video content, Zheng et al. incorporate a double-layered watermarking technique combined with blockchain technology~\cite{videoBlockchainWatermark}. The double watermarking technique embeds both robust and fragile watermarks into the video content, where the robust watermark ensures copyright protection, while the fragile watermark facilitates tamper detection and integrity verification. By integrating double watermarking with decentralization, the combined method achieved a 90\% precision rate and a 95\% recall rate in detecting tampered parts of watermarked videos against adversarial attacks. Sanivarapu et al. present a similarly robust digital watermarking system using cryptographic techniques to protect images from copyright infringement~\cite{digitalWatermarkCryptography}. The system first embeds a QR code watermark using Discrete Wavelet Transform (DWT)~\cite{dwtIntro}. It then layers on the transformed matrix with Singular Value Decomposition (SVD), and uses the RSA algorithm for watermark embedding, where the coefficients of DWT are modified using secret keys to embed the watermark, increasing the security and encryption of the image. While cryptographic watermarks provide better protection against specific threats like forgery and tampering, they also come with high computational overhead, and may be more perceptible than attention-based watermarks which can be optimized for processing through various generative models. 

% add pros and cons which is suitable for what area

\subsection{Cryptographic methods}
\vspace{0.4cm}
% add "other" in introduction (other than previous discussed watermarks focused appraoch, there's cryptographic specific method
Beyond applications in watermarks, advanced cryptographic strategies frequently aid in tracing and ownership verification of copyrighted content~\cite{Darwish2024,Shao2024}. One main type is digital signatures and hashing, which aim to authenticate content by providing a record of ownership and evidence of alterations or tampering that may be done to content, such as removing watermarks. Chain and Kuo (2013) use chaotic map transformations for generating digital signatures, using their unpredicable pattens to produce unique signatures from text for later verification~\cite{digSigChaotic}. By contrast, the Elliptic Curve Digital Signature Algorithm (ECDSA) used by Chandrashekhara et al. combines elliptic curve cryptography with digital signatures~\cite{compStudyDigSig}, using a SHA-256 hashing algorithm to generate a public key from a private one to authenticate the signature~\cite{verificationSHA256}. Commonly implemented along with hashing, blockchain methods similarly work to safeguard various works from infringement by recording transactions in cryptographically linked blocks, which are harder to tamper with than traditionally protected methods~\cite{researchBlockchainApp,blockchainMultiMedia}. The decentralized nature of blockchain enhances security by removing central points of control and creating a publicly accessible record of ownership. This can be particularly helpful for tracking data and model copyright status across each stage in the AI development life cycle, such as by integrating blockchain with contract management software for digital content~\cite{sai2024blockchaincontracts}.

\subsection{Deterring style transfer}
\vspace{0.4cm}
For image generation models, style transfer refers to the ability to produce the same content of a target image across a variety of artistic styles~\cite{imageStyleCNN}, which can lead to non-direct reproduction of copyrighted works. While style transfer is often explored as an intentional goal, such as for filling missing frames in an animation~\cite{Cai2023}, unintentional replication of an artist's style can increase copyright risks. Adversarial layers can help protect artistic work from being copied by adding a minimally perceptible layer that distorts the ability of an AI model to recognize it as a normal image and prevents the creation of derivative works~\cite{paintImitation, glaze, Zhong_2023}. For example, Li et al. further develop this strategy by using a momentum-based ensemble method to enhance the ability for these protections to be generalized across AI models~\cite{Li2024styletrans}. By altering the intermediate style representation of an image across multiple encoders and combining them through a softmax regression gradient, their method of ``Neural Style Protection'' can protect style transfer across both known and unknown models, while end-to-end and random noise baseline methods offer minimal protection.

\section{Preventing Copyright Infringement}
\vspace{0.4cm}
Moving on from methods to protect individual works, this section discusses prevention methods that AI developers can use to reduce the overall risk of a model generating reproductions of copyrighted works — a significant concern for developers seeking to avoid reputation loss and legal costs~\cite{Lucchi2023}. These methods focus on improving the general behavior of a model, and can be useful to apply at later stages in the AI development life cycle, including fine-tuning existing models. With many current datasets lacking proper content attribution or copyright information, these solutions may serve as a short-term way to enhance copyright compliance without needing to fully construct new large-scale datasets.

\subsection{Data de-duplication}
\vspace{0.4cm}
De-duplication refers to the process of removing redundant data within a model’s training dataset~\cite{comparativeDataDedup}, and has commonly been researched for the purposes of reducing data storage space~\cite{Zhou_2022_dedup} and improving query performance~\cite{RasinaBegum2021}. De-duplication methods may apply fingerprints or hash values to divided data chunks in order to reduce the computational resources needed to check for duplicates in a big data context~\cite{dedupBigCloud}. While existing research primarily focuses on the benefits of de-duplication for structured data and predictive models, Lee et al. show that de-duplication in generative AI data can improve overall performance and reduce verbatim copying in generative AI models~\cite{lee2022deduplication}. Particularly when combined with hashing or blockchain methods, de-duplication is effective at creating cleaner datasets for training generative AI and preventing copyright issues arising from the overuse of any particular material~\cite{practicalDedup, secureDedupBC}. De-duplication is especially well suited for cleaning ``noisy'' datasets, such as data scraped from social media~\cite{li2024gendedup}, as it can simultaneously reduce training time while improving performance and language understanding for LLMs.

\subsection{Regression and optimization}
\vspace{0.4cm}
In some cases, altering modeling and optimization choices can help prevent a model from reproducing copyrighted data, as Chu et al. ``copyright regression'' approach demonstrates~\cite{chu2023softmax}. By adding an inverse term to the training objective that discourages the model from generating outputs that match copyrighted data and demonstrating mathematically that it can be applied successfully on the softmax function, Chu et al. successfully create a method which helps balance between model performance and copyright protection. However, this method relies on knowledge about which training samples are copyrighted, which is not provided for most existing LLM datasets~\cite{renCopyrightProtection}. Kim et al. highlight the difficulty of applying an end-filter to remove copyrighted content, noting that no open source models are currently able to identify and differentiate copyrighted content, particularly as many datasets lack information about copyright attribution~\cite{kim2024jail}. While target datasets of copyrighted content have been created for the purposes of evaluating whether a model will produce them~\cite{kim2024jail}, filtering out all potential violations will require larger and more comprehensive datasets that contain clear information about attribution and copyright status of works. 

\subsection{Machine unlearning}
\vspace{0.3cm}
\subsubsection{Unlearning techniques}
\vspace{0.4cm}
For cases where an AI model has already been trained on copyrighted data, ``machine unlearning'' methods can be used to remove a group of samples from its training data, allowing it to act as though it has never seen the data before. This can be especially helpful from a legal perspective, allowing rules such as the GDPR's ``right to be forgotten"~\cite{chang2024forgotten} to be applied to machine learning models. Multiple methods have been developed to selectively remove content while preserving general features a model may have learned from the content. Zhang et al. introduce two different methods for unlearning: Elastic Weight Consolidation (EWC), which adds a constraint to the model's loss function to neutralize the influence of ``removed'' data, and Decreasing Moment Matching (DMM), which approximates the model's knowledge as a Gaussian distribution and selectively matches moments to similarly reduce reliance on data~\cite{machineUnlearnReversing}. 

\subsubsection{Generative AI specific strategies}
\vspace{0.4cm}
While earlier research mostly focused on unlearning for classification models, newer studies have explored unlearning for generative AI~\cite{liu2024rethink}. Liu et al. conduct a comprehensive survey of machine unlearning for generative models, categorizing them into two main approaches: parameter optimization, which focuses on adjusting model parameters linked with target removal data, and in-context unlearning, which alters input prompts through an API aiming to steer the model away from the ``unlearned'' content~\cite{liu2024unlearn}. 

\subsubsection{Knowledge entanglement}
\vspace{0.4cm}
One of the largest problems identified in both method types is knowledge entanglement, where data requiring removal is often closely tied to a model's knowledge of certain topics, causing a trade-off between model performance and compliance with the unlearning goal. To solve this issue, Tang et al. introduce a three-component framework to allow models to unlearn certain data without sacrificing their expressive capabilities~\cite{tang2024unlearning}. The three components include a Knowledge Unlearning Induction module which trains the model to forget specific sequences, a Contrastive Learning Enhancement module to maintain overall performance, and an Iterative Unlearning Refinement module to iteratively update the target data for unlearning, preventing a drastic shift to the model from ocurring. Similar strategies include adopting ``un-unlearning'' techniques to reintroduce unlearned data in context. This can be important to prevent the unlearned data from being recreated if introduced later as input to the system~\cite{shumailov2024ununlearning} or retained by association with similar concepts~\cite{kim2024jail}. For example, Van Gogh's \textit{Starry Night} may still be recreated after a model attempts to unlearn ``Van Gogh'' as it has high correlations with the concepts \textit{star} and \textit{night}. For this reason, unlearning strategies will often retain a copy of the unlearned data to serve as a reference for evaluating model outputs without being used to train the generative model~\cite{tang2024unlearning}. 

\section{Resources}
\vspace{0.3cm}
\subsection{Datasets}
\vspace{0.4cm}
While no comprehensive dataset of all copyrighted works has been developed~\cite{kim2024jail}, benchmark datasets of both AI-generated and human content serve as critical resources for testing and evaluating generative model performance on copyright issues. In this section, we present some of the most commonly used datasets for AI copyright research along with a discussion of their use potential. 

\subsubsection{Human content datasets} 
\vspace{0.4cm}
Repositories of human-generated content may serve as a benchmark for evaluating AI models by testing if an AI model completes a passage of protected text~\cite{langModelPlag} or comparing to AI-generated content for improving detection algorithms~\cite{yu2024aigc,he2024aigcvid,ImageAIGC}. Popular datasets for imagery include COCO~\cite{coco,microsoftCoco}, Flickr30K~\cite{flickr30k}, and OpenImages~\cite{openimages},  collectively offering over 9 million images with annotations describing the objects included and overall scene. Other datasets such as the the Metropolitan Museum of Art Open Access collection~\cite{theMetOpen} and WikiArt~\cite{hfWikiart} focus primarily on artistic contributions, which can be helpful for exploring style transfer or searching for matches between AI generations and human-created artwork. Text datasets include OpenSubtitles~\cite{opensubtitleDataset}, BookCorpus~\cite{bookcorpusDataset}, the WikiText collection from wikipedia articles~\cite{wikitextDataset}, and JSTOR's library of journal articles~\cite{jstorDataset} may be similarly used to scan for potential memorization, or compare human with AI-generated content. 

\subsubsection{Combined human-AI datasets} 
\vspace{0.4cm}
A separate category of datasets combines human-generated with AI-generated works across various art styles and writing subjects~\cite{aiArtBench,liyanage2022benchmark,gptWikiIntro}. These datasets are primarily valuable for training models to recognize text or image pairs, which allows for both detection and mitigation of copyright violations~\cite{deepCNN,dataDrivenModelReduction}. Some datasets in this category are designed for the specific purpose of comparing human copyrighted works to altered digital versions. Aboutalebi et al. introduce the Deepfake Art Challenge dataset, consisting of over 32,000 image pairs that are either forgeries, adversarially contaminated, or not~\cite{aboutalebi2024}. The selected methods used to modify images include inpainting, style transfer, adversarial data poisoning, and cutmix, representing popular copyright violation types such as modifying painting styles or using partial image data.

\nocite{wang2023fairness,chu2024history,saxena2023missed,zhang2019fairness,zhang2020flexible,zhang2020online,zhang2020learning,zhang2021farf,zhang2021fair,li2021time,zhang2023fairness,zhang2016using,zhang2018content,zhang2021autoencoder,zhang2018deterministic,tang2021interpretable,zhang2021disentangled,yazdanicomprehensive,liu2021research,liu2023segdroid,cai2023exploring,guyet2022incremental,zhang2024fairness}

\subsubsection{Feature / Artifact based datasets}
\vspace{0.4cm}
Beyond providing overall examples of human and AI creations, another category of datasets provides annotations over content to highlight specific `artifacts' in visual media, which can be used for both detecting and preventing violations. These `perceptual artifacts' capture subtle distortions or irregularities within an outputted image or video, which can be used for detecting the presence of AI in an image or ``inpainting'' to regenerate areas with potentially unwanted artifacts~\cite{deng2024defense}. Datasets such as PAL4Inpaint~\cite{zhang2022pal} and PAL4VST~\cite{zhang2023pal} can be used to localize artifacts within images which may show copyright violations, such as a distorted logo within an image. However, more research is necessary to understand how methods designed for deepfake detection and authentication may be applicable to copyright~\cite{deng2024defense}.

\subsection{Toolkits}
\vspace{0.3cm}
\subsubsection{Technical resources}
\vspace{0.4cm}
While most methods for detecting and mitigating AI copyright violations have yet to be applied extensively outside of a limited resource context, a few tools have been created for public use by AI developers and creators looking to protect their work. Copyright Catcher, developed by Patronus AI researchers, is the first API which aims to test for potential copyright violations in LLMs~\cite{patronus2024catcher}. Its method closely resembles those discussed in the evaluation methods subsection~\cite{kim2024jail,duarte2024decop}, using a dataset sampled from popular books on Goodreads to test if a model will complete the beginning of a prompt given from protected text. While this is a good start, it may not capture more complex copying behavior such as paraphrasing~\cite{langModelPlag}, and may have limited performance for models trained on other types of media. Most recently, researchers at  Imperial College London developed a ``copyright traps" system where creators can include ficticious entities in their content to detect where LLMs may be using their content, capturing behavior beyond pure memorization~\cite{meeus2024trap,brogan2024trap}. Resources have also been created to allow for better tracing and analysis of which data is typically used to train AI models. The Data Provenance Explorer, which allows practitioners to trace and filter on data provenance for the most popular open source data collections, is one such tool created for this purpose~\cite{longpre2023}. The interactive UI includes a tool to explore over 1800 popular datasets, viewing information such as their licenses, sources, and creators, along with additional visualizations about website protocols for scraping and AI training. However, this resource only focuses on text datasets, and more work is needed to establish provenance record tracing systems for other types of data. 

\subsubsection{Governance toolkits}
\vspace{0.4cm}
Several frameworks and toolkits have been developed for analyzing the ethical impacts of AI algorithms, such as the AI and data protection risk toolkit developed by the UK Information Commissioner's Office~\cite{icoTookit} and the Ethics \& Algorithms Toolkit developed in collaboration with the Center for Government Excellence at John Hopkins University~\cite{ethicsAlgToolkit}. However, no large-scale policy toolkits have been developed with the specific goal of increasing copyright compliance in AI. As legal and regulatory discussion around copyright in AI continues to evolve, one place where toolkits may play a helpful role is by enhancing public understanding of AI models in relation to copyright. Working to promote participatory approaches to AI and machine learning, Shen et al. established the Model Card Authoring Toolkit, a toolkit combining technical interfaces and collective decision-making protocols to empower community members to understand and make decisions about AI models in line with their collective values~\cite{shen2022card}. Applied to the issue of copyright, focus groups including authors, artists, and other rights holders could benefit from these tool kits to help understand and articulate their collective preferences surrounding AI models. 

\section{Challenges and Future Directions}
\vspace{0.4cm}
\sloppy
As researchers continue to work to understand and improve the behavior of generative AI models, several directions offer potential for discovering new information about identifying and mitigating copyright violations in generative AI. On the side of detection and evaluation, a persistent challenge is \textit{identifying indirect and ambiguous forms of copying} like paraphrasing or near-duplicate code generation~\cite{copybench,ducru2024royalties,xu2024license}. More refined benchmarks and detection tools are needed to improve copyright enforcement for these cases. In a similar vein, \textit{building copyright-specific datasets} is necessary to facilitate better research and detection of copyright issues at scale, track content ownership, and help facilitate certain technical mitigation strategies. For technical methods protecting copyrighted works, \textit{strengthening watermark resilience} remains a major challenge, as there is often a trade-off between imperceptibility and security, and common modifications like compression or noise can easily weaken or remove existing markers~\cite{zhao2024advwatermark}. On the side of preventing copyright infringement, more research is needed to understand how AI model parameters can reduce copyright violations while \textit{retaining maximum utility} and \textit{propagating to downstream models} which may be fine-tuned for different goals. Furthermore, there is a need to explore how domain-specific methods can be \textit{transferrable across multiple types of content}, as many current strategies are limited in scope, which may hamper coordination and research. Lastly, each of these efforts continually adapt as the legal system around generative AI evolves. Stronger \textit{consensus on the bounds of fair use} can help strengthen copyright enforcement and bridge the interests of IP rights holders and AI developers. 

%\wenbin{conclusion does not contain citation nor abstract. -- Done}

\section{Conclusion}

\vspace{0.4cm}

As generative AI systems trained on copyrighted data continue to proliferate in the absence of clear legal frameworks, a combination of technical and policy measures is necessary to prevent copyright infringement and ensure that generative AI is developed with fair use principles in mind. Creators, developers, policymakers, and other stakeholders should take action to carefully assess their current compliance with copyright standards and establish stronger systems of oversight, mitigation, and accountability for copyright harms caused by AI. In many cases, this will involve a continued dialogue between rights holders and AI providers to understand their demands and increase transparency surrounding technical design choices. Researchers should also expand efforts to detect violations, protect creative works, and improve the copyright performance of AI models to address the challenges previously mentioned. Through this integrated approach, creators, developers, and policymakers can collaborate to promote an AI ecosystem designed to support fair use and foster human creativity.

% \newpage

\sloppy

\newpage

\section{REFERENCES}
\vspace{0.3cm}

\printbibliography[heading=none]

@inproceedings{wang2023preventing,
	title={Preventing Discriminatory Decision-making in Evolving Data Streams},
	author={Wang, Zichong and Saxena, Nripsuta and Yu, Tongjia and Karki, Sneha and Zetty, Tyler and Haque, Israat and Zhou, Shan and Kc, Dukka and Stockwell, Ian and Bifet, Albert and others},
	booktitle={Proceedings of the 2023 ACM Conference on Fairness, Accountability, and Transparency (FAccT)},
	year={2023}
}

@inproceedings{zhang2023individual,
	title={Individual Fairness under Uncertainty},
	author={Zhang, Wenbin and Wang, Zichong and Kim, Juyong and Cheng, Chen and Oommen, Thomas and Ravikumar, Pradeep and Weiss, Jeremy},
	booktitle={26th European Conference on Artificial Intelligence},
	pages={3042--3049},
	year={2023}
}

@inproceedings{wang2023fg2an,
	title={FG$^2$AN: Fairness-Aware Graph Generative Adversarial Networks},
	author={Wang, Zichong and Wallace, Charles and Bifet, Albert and Yao, Xin and Zhang, Wenbin},
	booktitle={Joint European Conference on Machine Learning and Knowledge Discovery in Databases},
	pages={259--275},
	year={2023},
	organization={Springer Nature Switzerland}
}

@article{yazdani2024comprehensive,
	title={A Comprehensive Survey of Image and Video Generative AI: Recent Advances, Variants, and Applications},
	author={Yazdani, Shamim and Saxena, Nripsuta and Wang, Zichong and Wu, Yanzhao and Zhang, Wenbin},
	year={2024}
}

@inproceedings{wang2023mitigating,
	title={Mitigating multisource biases in graph neural networks via real counterfactual samples},
	author={Wang, Zichong and Narasimhan, Giri and Yao, Xin and Zhang, Wenbin},
	booktitle={2023 IEEE International Conference on Data Mining (ICDM)},
	pages={638--647},
	year={2023},
	organization={IEEE}
}

@inproceedings{chinta2023optimization,
	title={Optimization and improvement of fake news detection using voting technique for societal benefit},
	author={Chinta, Sribala Vidyadhari and Fernandes, Karen and Cheng, Ningxi and Fernandez, Jordan and Yazdani, Shamim and Yin, Zhipeng and Wang, Zichong and Wang, Xuyu and Xu, Weifeng and Liu, Jun and others},
	booktitle={2023 IEEE International Conference on Data Mining Workshops (ICDMW)},
	pages={1565--1574},
	year={2023},
	organization={IEEE}
}

@article{wang2024history,
	title={History, Development, and Principles of Large Language Models-An Introductory Survey},
	author={Wang, Zichong and Chu, Zhibo and Doan, Thang Viet and Ni, Shiwen and Yang, Min and Zhang, Wenbin},
	journal={AI and Ethics, 2024},
	year={2024},
	publisher={Springer}
}

@article{chu2024fairness,
	title={Fairness in Large Language Models: A Taxonomic Survey},
	author={Chu, Zhibo and Wang, Zichong and Zhang, Wenbin},
	journal={ACM SIGKDD Explorations Newsletter, 2024},
	pages={34--48},
	year={2024}
}

@article{dzuong2024uncertain,
	title={Uncertain Boundaries: Multidisciplinary Approaches to Copyright Issues in Generative AI},
	author={Dzuong, Jocelyn and Wang, Zichong and Zhang, Wenbin},
	journal={arXiv preprint arXiv:2404.08221},
	year={2024}
}

@inproceedings{yin2024improving,
	title={Improving Fairness in Machine Learning Software via Counterfactual Fairness Thinking},
	author={Yin, Zhipeng and Wang, Zichong and Zhang, Wenbin},
	booktitle={Proceedings of the 2024 IEEE/ACM 46th International Conference on Software Engineering: Companion Proceedings},
	pages={420--421},
	year={2024}
}

@article{wang2023towards,
	title={Towards fair machine learning software: Understanding and addressing model bias through counterfactual thinking},
	author={Wang, Zichong and Zhou, Yang and Haque, Israat and Lo, David and Zhang, Wenbin},
	journal={arXiv preprint arXiv:2302.08018},
	year={2023}
}

@article{wang2024toward,
	title={Toward Fair Graph Neural Networks via Real Counterfactual Samples},
	author={Wang, Zichong and Qiu, Meikang and Chen, Min and Salem, Malek Ben and Yao, Xin and Zhang, Wenbin},
	journal={Knowledge and Information Systems},
	pages={1--25},
	year={2024},
	publisher={Springer London}
}

@article{chinta2024fairaied,
	title={FairAIED: Navigating fairness, bias, and ethics in educational AI applications},
	author={Chinta, Sribala Vidyadhari and Wang, Zichong and Yin, Zhipeng and Hoang, Nhat and Gonzalez, Matthew and Quy, Tai Le and Zhang, Wenbin},
	journal={arXiv preprint arXiv:2407.18745},
	year={2024}
}

@article{doan2024fairness1,
	title={Fairness definitions in language models explained},
	author={Wang, Zichong and Palikhe, Avash and Yin, Zhipeng and Zhang, Wenbin},
	journal={arXiv preprint arXiv:2407.18454},
	year={2024}
}

@inproceedings{wang2024individual,
	title={Individual Fairness with Group Awareness Under Uncertainty},
	author={Wang, Zichong and Dzuong, Jocelyn and Yuan, Xiaoyong and Chen, Zhong and Wu, Yanzhao and Yao, Xin and Zhang, Wenbin},
	booktitle={Joint European Conference on Machine Learning and Knowledge Discovery in Databases},
	pages={89--106},
	year={2024},
	organization={Springer Nature Switzerland}
}

@inproceedings{doan2024fairness,
	title={Fairness in large language models in three hours},
	author={Doan, Thang Viet and Wang, Zichong and Hoang, Nhat Nguyen Minh and Zhang, Wenbin},
	booktitle={Proceedings of the 33rd ACM International Conference on Information and Knowledge Management},
	pages={5514--5517},
	year={2024}
}

@inproceedings{wang2024advancing,
	title={Advancing Graph Counterfactual Fairness through Fair Representation Learning},
	author={Wang, Zichong and Chu, Zhibo and Blanco, Ronald and Chen, Zhong and Chen, Shu-Ching and Zhang, Wenbin},
	booktitle={Joint European Conference on Machine Learning and Knowledge Discovery in Databases},
	pages={40--58},
	year={2024},
	organization={Springer Nature Switzerland}
}

@inproceedings{wang2024group,
	title={Group Fairness with Individual and Censorship Constraints},
	author={Wang, Zichong and Zhang, Wenbin},
	booktitle={27th European Conference on Artificial Intelligence},
	year={2024}
}

@inproceedings{yin2024accessible,
	title={Accessible Health Screening Using Body Fat Estimation by Image Segmentation},
	author={Yin, Zhipeng and Agarwal, Sameeksha and Kashif, Ayesha and Gonzalez, Matthew and Wang, Zichong and Liu, Suqing and Liu, Zhen and Wu, Yanzhao and Stockwell, Ian and Xu, Weifeng and others},
	booktitle={2024 IEEE International Conference on Data Mining Workshops (ICDMW)},
	pages={405--414},
	year={2024}
}

@article{wang2025fg,
	title={FG-SMOTE: Towards Fair Node Classification with Graph Neural Network},
	author={Wang, Zichong and Yin, Zhipeng and Zhang, Yuying and Yang, Liping and Zhang, Tingting and Pissinou, Niki and Cai, Yu and Hu, Shu and Li, Yun and Zhao, Liang and others},
	journal={ACM SIGKDD Explorations Newsletter},
	volume={26},
	number={2},
	pages={99--108},
	year={2025},
	publisher={ACM New York, NY, USA}
}

@article{wang2025graph,
	title={Graph Fairness via Authentic Counterfactuals: Tackling Structural and Causal Challenges},
	author={Wang, Zichong and Yin, Zhipeng and Liu, Fang and Liu, Zhen and Lisetti, Christine and Yu, Rui and Wang, Shaowei and Liu, Jun and Ganapati, Sukumar and Zhou, Shuigeng and others},
	journal={ACM SIGKDD Explorations Newsletter},
	volume={26},
	number={2},
	pages={89--98},
	year={2025},
	publisher={ACM}
}

@inproceedings{wang2025fair,
	title={Fair Graph U-Net: A Fair Graph Learning Framework Integrating Group and Individual Awareness},
	author={Wang, Zichong and Chu, Zhibo and Viet Doan, Thang and Wang, Shaowei and Wu, Yongkai and Palade, Vasile and Zhang, Wenbin},
	booktitle={proceedings of the AAAI conference on artificial intelligence},
	volume={39},
	number={27},
	pages={28485--28493},
	year={2025}
}

@inproceedings{wang2025towards,
	title={Towards Fair Graph Learning without Demographic Information},
	author={Wang, Zichong and Hoang, Nhat and Zhang, Xingyu and Bello, Kevin and Zhang, Xiangliang and Iyengar, Sundararaja Sitharama and Zhang, Wenbin},
	booktitle={The 28th International Conference on Artificial Intelligence and Statistics},
	volume={258},
	pages={2107--2115},
	year={2025}
}

@article{chu2024history,
  title={History, Development, and Principles of Large Language Models-An Introductory Survey},
  author={Chu, Zhibo and Ni, Shiwen and Wang, Zichong and Feng, Xi and Li, Chengming and Hu, Xiping and Xu, Ruifeng and Yang, Min and Zhang, Wenbin},
  journal={arXiv preprint arXiv:2402.06853},
  year={2024}
}

@inproceedings{saxena2023missed,
  title={Missed Opportunities in Fair AI},
  author={Saxena, Nripsuta Ani and Zhang, Wenbin and Shahabi, Cyrus},
  booktitle={Proceedings of the 2023 SIAM International Conference on Data Mining (SDM)},
  pages={961--964},
  year={2023},
  organization={SIAM}
}

@article{yin2025digital,
	title={Digital Forensics in the Age of Large Language Models},
	author={Yin, Zhipeng and Wang, Zichong and Xu, Weifeng and Zhuang, Jun and Mozumder, Pallab and Smith, Antoinette and Zhang, Wenbin},
	journal={arXiv preprint arXiv:2504.02963},
	year={2025}
}

@article{chinta2025ai,
	title={AI-driven healthcare: Fairness in AI healthcare: A survey},
	author={Chinta, Sribala Vidyadhari and Wang, Zichong and Palikhe, Avash and Zhang, Xingyu and Kashif, Ayesha and Smith, Monique Antoinette and Liu, Jun and Zhang, Wenbin},
	journal={PLOS Digital Health},
	volume={4},
	number={5},
	pages={e0000864},
	year={2025},
	publisher={Public Library of Science San Francisco, CA USA}
}

@inproceedings{wang2025fdgen,
	title     = {FDGen: A Fairness-Aware Graph Generation Model},
	author    = {Wang, Zichong and Zhang, Wenbin},
	booktitle = {Proceedings of the 42nd International Conference on Machine Learning},
	year      = {2025},
	organization={PMLR}
}

@inproceedings{wang2025fairgnn,
	title     = {fairGNN-WOD: Fair Graph Learning Without Complete Demographics},
	author    = {Wang, Zichong and Liu, Fang and Pan, Shimei and Liu, Jun and Saeed, Fahad and Qiu, Meikang and Zhang, Wenbin},
	booktitle = {Proceedings of the 34th International Joint Conference on Artificial Intelligence},
	year      = {2025}
}

@inproceedings{wang2025limiteddemographics,
	author    = {Wang, Zichong and Wu, Anqi and Moniz, Nuno and Hu, Shu and Knijnenburg, Bart and Zhu, Qingquan and Zhang, Wenbin},
	title     = {Towards Fairness with Limited Demographics via Disentangled Learning},
	booktitle = {Proceedings of the 34th International Joint Conference on Artificial Intelligence},
	year      = {2025}
}

@inproceedings{wang2025Fairness,
	title={Fairness-Aware Graph Representation Learning Without Demographic Information},
	author={Wang, Zichong and Yin, Zhipeng and Yang, Liping and Zhuang, Jun and Yu, Rui and Kong,Qingzhao and Zhang, Wenbin},
	booktitle={Joint European Conference on Machine Learning and Knowledge Discovery in Databases},
	year={2025},
	organization={Springer Nature Switzerland}
}

@inproceedings{wang2025Redefining,
	title={Redefining Fairness: A Multi-dimensional Perspective and Integrated Evaluation Framework},
	author={Wang, Zichong and Yin, Zhipeng and Yap, Roland and Zhang, Xiaocai and Hu, Shu and Zhang, Wenbin},
	booktitle={Joint European Conference on Machine Learning and Knowledge Discovery in Databases},
	year={2025},
	organization={Springer Nature Switzerland}
}

@inproceedings{zhang2019fairness,
	title={On fairness-aware learning for non-discriminative decision-making},
	author={Zhang, Wenbin and Tang, Xuejiao and Wang, Jianwu},
	booktitle={International Conference on Data Mining Workshops (ICDMW)},
	pages={1072--1079},
	year={2019}
}

@inproceedings{zhang2020flexible,
	title={Flexible and adaptive fairness-aware learning in non-stationary data streams},
	author={Zhang, Wenbin and others},
	booktitle={IEEE 32nd International Conference on Tools with Artificial Intelligence (ICTAI)},
	pages={399--406},
	year={2020}
}

@inproceedings{zhang2021fair,
	title={Fair Decision-making Under Uncertainty},
	author={Zhang, Wenbin and Weiss, Jeremy},
	booktitle={{2021 IEEE International Conference on Data Mining (ICDM)}},
	year={2021},
	organization={IEEE}
}

@article{zhang2022fairness,
	title={Fairness Amidst Non-IID Graph Data: Current Achievements and Future Directions},
	author={Zhang, Wenbin and Pan, Shimei and Zhou, Shuigeng and Walsh, Toby and Weiss, Jeremy C},
	journal={arXiv preprint arXiv:2202.07170},
	year={2022}
}

@inproceedings{zhang2021farf,
	title={FARF: A Fair and Adaptive Random Forests Classifier},
	author={Zhang, Wenbin and Bifet, Albert and Zhang, Xiangliang and Weiss, Jeremy C and Nejdl, Wolfgang},
	booktitle={Pacific-Asia Conference on Knowledge Discovery and Data Mining},
	pages={245--256},
	year={2021},
	organization={Springer}
}

@article{zhang2023fairness,
  title={Fairness with censorship and group constraints},
  author={Zhang, Wenbin and Weiss, Jeremy C},
  journal={Knowledge and Information Systems},
  pages={1--24},
  year={2023},
  publisher={Springer}
}

@inproceedings{zhang2016using,
	title={Using the machine learning approach to predict patient survival from high-dimensional survival data},
	author={Zhang, Wenbin and Tang, Jian and Wang, Nuo},
	booktitle={IEEE International Conference on Bioinformatics and Biomedicine (BIBM)},
	year={2016}
}

@inproceedings{zhang2018content,
	title={Content-bootstrapped collaborative filtering for medical article recommendations},
	author={Zhang, Wenbin and Wang, Jianwu},
	booktitle={IEEE International Conference on Bioinformatics and Biomedicine (BIBM)},
	year={2018}
}

@inproceedings{zhang2018deterministic,
	title={A deterministic self-organizing map approach and its application on satellite data based cloud type classification},
	author={Zhang, Wenbin and Wang, Jianwu and Jin, Daeho and Oreopoulos, Lazaros and Zhang, Zhibo},
	booktitle={IEEE International Conference on Big Data (Big Data)},
	year={2018}
}

@inproceedings{tang2021interpretable,
  title={Interpretable Visual Understanding with Cognitive Attention Network},
  author={Tang, Xuejiao and Zhang, Wenbin and Yu, Yi and Turner, Kea and Derr, Tyler and Wang, Mengyu and Ntoutsi, Eirini},
  booktitle={International Conference on Artificial Neural Networks},
  pages={555--568},
  year={2021},
  organization={Springer}
}

@inproceedings{zhang2021autoencoder,
  title={Autoencoder for neuroimage},
  author={Zhang, Mingli and Zhang, Fan and Zhang, Jianxin and Chaddad, Ahmad and Guo, Fenghua and Zhang, Wenbin and Zhang, Ji and Evans, Alan},
  booktitle={International conference on database and expert systems applications},
  pages={84--90},
  year={2021},
  organization={Springer}
}

@article{zhang2020learning,
	title={Learning fairness and graph deep generation in dynamic environments},
	author={Zhang, Wenbin},
	year={2020},
}

@inproceedings{zhang2021disentangled,
	title={Disentangled Dynamic Graph Deep Generation},
	author={Zhang, Wenbin and Zhang, Liming and Pfoser, Dieter and Zhao, Liang},
	booktitle={Proceedings of the SIAM International Conference on Data Mining (SDM)},
	pages={738--746},
	year={2021}
}

@article{yazdanicomprehensive,
  title={A Comprehensive Survey of Image and Video Generative AI: Recent Advances, Variants, and Applications},
  author={Yazdani, Shamim and Saxena, Nripsuta and Wang, Zichong and Wu, Yanzhao and Zhang, Wenbin},
  year={2024}
}

@article{liu2021research,
  title={Research on unsupervised feature learning for Android malware detection based on Restricted Boltzmann Machines},
  author={Liu, Zhen and Wang, Ruoyu and Japkowicz, Nathalie and Tang, Deyu and Zhang, Wenbin and Zhao, Jie},
  journal={Future Generation Computer Systems},
  volume={120},
  pages={91--108},
  year={2021},
  publisher={Elsevier}
}

@inproceedings{cai2023exploring,
  title={Exploring Approaches for Teaching Cybersecurity and AI for K-12},
  author={Cai, Yu and Youngstrom, Drew and Zhang, Wenbin},
  booktitle={2023 IEEE International Conference on Data Mining Workshops (ICDMW)},
  pages={1559--1564},
  year={2023},
  organization={IEEE}
}

@inproceedings{guyet2022incremental,
  title={Incremental Mining of Frequent Serial Episodes Considering Multiple Occurrences},
  author={Guyet, Thomas and Zhang, Wenbin and Bifet, Albert},
  booktitle={22nd International Conference on Computational Science},
  pages={460--472},
  year={2022},
  organization={Springer}
}

@inproceedings{zhang2024fairness,
  title={Fairness with Censorship: Bridging the Gap between Fairness Research and Real-World Deployment},
  author={Zhang, Wenbin},
  booktitle={Proceedings of the AAAI Conference on Artificial Intelligence},
  volume={38},
  number={20},
  pages={22685--22685},
  year={2024}
}

@article{liu2023segdroid,
  title={SeGDroid: An Android malware detection method based on sensitive function call graph learning},
  author={Liu, Zhen and Wang, Ruoyu and Japkowicz, Nathalie and Gomes, Heitor Murilo and Peng, Bitao and Zhang, Wenbin},
  journal={Expert Systems with Applications},
  pages={121125},
  year={2023},
  publisher={Elsevier}
}

@article{zhang2020online,
	title={Online decision trees with fairness},
	author={Zhang, Wenbin and Zhao, Liang},
	journal={arXiv preprint arXiv:2010.08146},
	year={2020}
}

@inproceedings{zhang2019faht,
	title={FAHT: an adaptive fairness-aware decision tree classifier},
	author={Zhang, Wenbin and Ntoutsi, Eirini},
	booktitle={Proceedings of the 28th International Joint Conference on Artificial Intelligence},
	pages={1480--1486},
	year={2019}
}

@article{zhang2024ai,
	title={AI fairness in practice: Paradigm, challenges, and prospects},
	author={Zhang, Wenbin},
	journal={Ai Magazine},
	volume={45},
	number={3},
	pages={386--395},
	year={2024},
	publisher={Wiley Online Library}
}

@inproceedings{zhang2022longitudinal,
	title={Longitudinal fairness with censorship},
	author={Zhang, Wenbin and Weiss, Jeremy C},
	booktitle={proceedings of the AAAI conference on artificial intelligence},
	volume={36},
	number={11},
	pages={12235--12243},
	year={2022}
}

@inproceedings{zhang2023censored,
	title={Censored fairness through awareness},
	author={Zhang, Wenbin and Hernandez-Boussard, Tina and Weiss, Jeremy},
	booktitle={Proceedings of the AAAI conference on artificial intelligence},
	volume={37},
	number={12},
	pages={14611--14619},
	year={2023}
}

@article{zhang2025fairness,
	title={Fairness amidst non-IID graph data: A literature review},
	author={Zhang, Wenbin and Zhou, Shuigeng and Walsh, Toby and Weiss, Jeremy C},
	journal={AI Magazine},
	volume={46},
	number={1},
	pages={e12212},
	year={2025},
	publisher={Wiley Online Library}
}

@article{Elkhatat2023,
  title = {Evaluating the efficacy of AI content detection tools in differentiating between human and AI-generated text},
  volume = {19},
  ISSN = {1833-2595},
  url = {http://dx.doi.org/10.1007/s40979-023-00140-5},
  DOI = {10.1007/s40979-023-00140-5},
  number = {1},
  journal = {International Journal for Educational Integrity},
  publisher = {Springer Science and Business Media LLC},
  author = {Elkhatat,  Ahmed M. and Elsaid,  Khaled and Almeer,  Saeed},
  year = {2023},
  month = sep 
}

@misc{huang2024ruai,
      title={RU-AI: A Large Multimodal Dataset for Machine Generated Content Detection}, 
      author={Liting Huang and Zhihao Zhang and Yiran Zhang and Xiyue Zhou and Shoujin Wang},
      year={2024},
      eprint={2406.04906},
      archivePrefix={arXiv},
      primaryClass={cs.CV},
      url={https://arxiv.org/abs/2406.04906}, 
}

@ARTICLE{ImageAIGC,
  author={Park, Daeeol and Na, Hyunsik and Choi, Daeseon},
  journal={IEEE Access}, 
  title={Performance Comparison and Visualization of AI-Generated-Image Detection Methods}, 
  year={2024},
  volume={12},
  number={},
  pages={62609-62627},
  doi={10.1109/ACCESS.2024.3394250}}

@misc{yu2024aigc,
      title={Fake Artificial Intelligence Generated Contents (FAIGC): A Survey of Theories, Detection Methods, and Opportunities}, 
      author={Xiaomin Yu and Yezhaohui Wang and Yanfang Chen and Zhen Tao and Dinghao Xi and Shichao Song and Simin Niu and Zhiyu Li},
      year={2024},
      eprint={2405.00711},
      archivePrefix={arXiv},
      primaryClass={cs.CL},
      url={https://arxiv.org/abs/2405.00711}, 
}

@misc{he2024aigcvid,
      title={Exposing AI-generated Videos: A Benchmark Dataset and a Local-and-Global Temporal Defect Based Detection Method}, 
      author={Peisong He and Leyao Zhu and Jiaxing Li and Shiqi Wang and Haoliang Li},
      year={2024},
      eprint={2405.04133},
      archivePrefix={arXiv},
      primaryClass={cs.CV},
      url={https://arxiv.org/abs/2405.04133}, 
}

@article{renCopyrightProtection,
    author = {Ren, Jie and Xu, Han and He, Pengfei and Cui, Yingqian and Zeng, Shenglai and Zhang, Jiankun and Wen, Hongzhi and Ding, Jiayuan and Liu, Hui and Chang, Yi and Tang, Jiliang},
    title = {Copyright Protection in Generative AI: A Technical Perspective},
    journal = {arXiv},
    year = {2024},
    url = {https://arxiv.org/abs/2402.02333}
}

@article{li2021blackbox,
  title = {Does black-box machine learning shift the US fair use doctrine?},
  volume = {16},
  ISSN = {1747-1540},
  url = {http://dx.doi.org/10.1093/jiplp/jpab118},
  DOI = {10.1093/jiplp/jpab118},
  number = {11},
  journal = {Journal of Intellectual Property Law \& Practice},
  publisher = {Oxford University Press (OUP)},
  author = {Li,  Yangzi},
  year = {2021},
  month = sep,
  pages = {1175–1185}
}

@inproceedings{glaze,
author = {Shan, Shawn and Cryan, Jenna and Wenger, Emily and Zheng, Haitao and Hanocka, Rana and Zhao, Ben Y.},
title = {Glaze: protecting artists from style mimicry by text-to-image models},
year = {2023},
isbn = {978-1-939133-37-3},
publisher = {USENIX Association},
address = {USA},
booktitle = {Proceedings of the 32nd USENIX Conference on Security Symposium},
articleno = {123},
numpages = {18},
location = {Anaheim, CA, USA},
series = {SEC '23}
}

@inproceedings{secureDedupBC,
    author = {Aparna, R. and Kulkarni, Roopa G. and Chaudhari, Shilpa},
    title = {Secure Deduplication for Images using Blockchain},
    booktitle = {IEEE International Conference on Electronics, Computing and Communication Technologies (CONECCT)},
    year = {2020}
}

@article{diffusionShield,
    author = {Cui, Yingqian and Ren, Jie and Xu, Han and He, Pengfei and Liu, Hui and Sun, Lichao and Xing, Yue and Tang, Jiliang},
    title = {DiffusionShield: A Watermark for Copyright Protection Against Generative Diffusion Models},
    journal = {NeurIPS 2023 Workshop on Diffusion Models},
    year = {2023}
}

@article{paintImitation,
    author = {Liang, Chumeng and Wu, Xiaoyu and Hua, Yang and Zhang, Jiaru and Xue, Yiming and Song, Tao and Xue, Zhengui and Ma, Ruhui and Guan, Haibing},
    publisher = {2023 International Conference on Machine Learning},
    title = {Adversarial Example Does Good: Preventing Painting Imitation from Diffusion Models via Adversarial Examples},
    year = {2023}
}

@misc{Stempel_2023, title={NY Times sues openai, Microsoft for infringing copyrighted works ...}, url={https://www.reuters.com/legal/transactional/ny-times-sues-openai-microsoft-infringing-copyrighted-work-2023-12-27/}, journal={Reuters}, publisher={Thomson Reuters Corporation}, author={Stempel, Jonathan}, year={2023}, month={Dec}}

@misc{defintion-copyright-office, title={Definitions (FAQ) | U.S. Copyright Office}, url={https://www.copyright.gov/help/faq/faq-definitions.html}, journal={Copyright.gov}, author={U.S. Copyright Office}, year={2008} }

@article{thompson_2024, title={We Asked A.I. to Create the Joker. It Generated a Copyrighted Image.}, url={https://www.nytimes.com/interactive/2024/01/25/business/ai-image-generators-openai-microsoft-midjourney-copyright.html}, journal={The New York Times}, author={Thompson, Stuart A.}, year={2024}, month={Jan} }

@misc{visPlagiarism, title={Generative AI Has a Visual Plagiarism Problem - IEEE Spectrum}, url={https://spectrum.ieee.org/midjourney-copyright}, journal={IEEE Spectrum}, publisher={Institute of Electrical and Electronics Engineers (IEEE)}, author={Marcus, Gary and Southen, Reid}, year={2024}, month={Jan} }

@misc{midjourneyTerms, title={Midjourney Terms of Service}, url={https://docs.midjourney.com/docs/terms-of-service}, journal={Midjourney}, author={Midjourney}, year={2023}, month={Feb} }

@article{robustVidWatermark, title={Robust Invisible Video Watermarking with Attention.}, url={https://arxiv.org/abs/1909.01285}, author={Zhang, Kevin A. and Xu, Lei and Cuesta-Infante, Alfredo and Kalyan Veeramachaneni}, year={2019}, month={Sep} }

@article{videoBlockchainWatermark, title={A Novel Video Copyright Protection Scheme Based on Blockchain and Double Watermarking}, volume={2021}, DOI={https://doi.org/10.1155/2021/6493306}, journal={Security and Communication Networks}, author={Zheng, Jingjing and Teng, Shuhua and Li, Peirong and Ou, Wei and Zhou, Donghao and Ye, Jun}, editor={Gao, Honghao}, year={2021}, month={Dec}, pages={1–16} }

@article{tchebichefWatermarking, title={An Improved Watermarking Technique for Copyright Protection Based on Tchebichef Moments}, volume={7}, DOI={https://doi.org/10.1109/access.2019.2948086}, journal={IEEE Access}, author={Ernawan, Ferda and Kabir, Muhammad Nomani}, year={2019}, pages={151985–152003} }

@article{imageTchebichef,
  author={Mukundan, R. and Ong, S.H. and Lee, P.A.},
  journal={IEEE Transactions on Image Processing}, 
  title={Image analysis by Tchebichef moments}, 
  year={2001},
  volume={10},
  number={9},
  pages={1357-1364},
  doi={10.1109/83.941859}}

@article{digitalWatermarkCryptography, title={Digital Watermarking System for Copyright Protection and Authentication of Images Using Cryptographic Techniques}, volume={12}, DOI={https://doi.org/10.3390/app12178724}, number={17}, journal={Applied Sciences}, author={Sanivarapu, Prasanth Vaidya and Rajesh, Kandala N. V. P. S. and Hosny, Khalid M. and Fouda, Mostafa M.}, year={2022}, month={Aug}, pages={1-13} }

@article{decentralizedFingerprinting, title={Survey on Decentralized Fingerprinting Solutions: Copyright Protection through Piracy Tracing}, volume={9}, DOI={https://doi.org/10.3390/computers9020026}, number={2}, journal={Computers}, author={Megías, David and Kuribayashi, Minoru and Qureshi, Amna}, year={2020}, month={Apr}, pages={1-19} }

@book{dwtIntro, title={Discrete Wavelet Transform: an Introduction}, author={Skodras, Athanassios}, year={2003}, month={Feb}, pages={1–26} }

@article{machineUnlearnReversing, title={Machine Unlearning by Reversing the Continual Learning}, volume={13}, DOI={https://doi.org/10.3390/app13169341}, number={16}, journal={Applied Sciences}, publisher={Multidisciplinary Digital Publishing Institute}, author={Zhang, Yongjing and Lu, Zhaobo and Zhang, Feng and Wang, Hao and Li, Shaojing}, year={2023}, month={Aug}, pages={1-10} }

@article{dataDrivenModelReduction,
title = {Data-driven model reduction by moment matching for linear and nonlinear systems},
journal = {Automatica},
volume = {79},
pages = {340-351},
year = {2017},
issn = {0005-1098},
doi = {https://doi.org/10.1016/j.automatica.2017.01.014},
url = {https://www.sciencedirect.com/science/article/pii/S0005109817300249},
author = {Scarciotti, Giordano and Astolfi, Alessandro}
}

@inproceedings{comparativeDataDedup,
  author={Chhabra, Nipun and Bala, Manju},
  booktitle={2018 First International Conference on Secure Cyber Computing and Communication (ICSCCC)}, 
  title={A Comparative Study of Data Deduplication Strategies}, 
  year={2018},
  pages={68-72},
  doi={10.1109/ICSCCC.2018.8703363}}

@article{practicalDedup, title={A study of practical deduplication}, volume={7}, DOI={https://doi.org/10.1145/2078861.2078864}, number={4}, journal={ACM Transactions on Storage}, author={Meyer, Dutch T. and Bolosky, William J.}, year={2012}, month={Jan}, pages={1–20} }

@inproceedings{imageStyleCNN,
  author={Gatys, Leon A. and Ecker, Alexander S. and Bethge, Matthias},
  booktitle={2016 IEEE Conference on Computer Vision and Pattern Recognition (CVPR)}, 
  title={Image Style Transfer Using Convolutional Neural Networks}, 
  year={2016},
  pages={2414-2423},
  doi={10.1109/CVPR.2016.265}}

@article{digSigChaotic,
author = {Chain, Kai and Kuo, Wen-Chung},
year = {2013},
month = {12},
title = {A new digital signature scheme based on chaotic maps},
volume = {74},
journal = {Nonlinear Dynamics},
doi = {10.1007/s11071-013-1018-1}
}

@article{compStudyDigSig, title={A COMPREHENSIVE STUDY ON DIGITAL SIGNATURE}, volume={9}, DOI={https://doi.org/10.21276/ijircst.2021.9.3.7}, number={3}, journal={International Journal of Innovative Research in Computer Science \& Technology}, author={Chandrashekhara, J. and V B, Anu and H, Prabhavathi and B R, Ramya}, year={2021}, month={May} }

@article{verificationSHA256,
author = {Appel, Andrew W.},
title = {Verification of a Cryptographic Primitive: SHA-256},
year = {2015},
issue_date = {April 2015},
journal = {ACM Transactions on Programming Languages and Systems},
address = {New York, NY, USA},
volume = {37},
number = {2},
issn = {0164-0925},
url = {https://doi.org/10.1145/2701415},
doi = {10.1145/2701415},
month = {apr},
articleno = {7},
numpages = {31}
}

@inproceedings{researchBlockchainApp,
  author={Jiang, Tao and Sui, Aina and Lin, Weiguo and Han, Pengbin},
  booktitle={2020 International Conference on Culture-oriented Science \& Technology (ICCST)}, 
  title={Research on the Application of Blockchain in Copyright Protection}, 
  year={2020},
  pages={616-621},
  doi={10.1109/ICCST50977.2020.00127}}

@Article{blockchainMultiMedia,
AUTHOR = {Qureshi, Amna and Megías Jiménez, David},
TITLE = {Blockchain-Based Multimedia Content Protection: Review and Open Challenges},
JOURNAL = {Applied Sciences},
VOLUME = {11},
YEAR = {2021},
NUMBER = {1},
ARTICLE-NUMBER = {1},
URL = {https://www.mdpi.com/2076-3417/11/1/1},
ISSN = {2076-3417},
DOI = {10.3390/app11010001}
}

@misc{aiArtBench, title={AI-ArtBench}, url={https://www.kaggle.com/datasets/ravidussilva/real-ai-art}, journal={Kaggle}, author={Silva, Ravidu and Bird, Jordan J.}, year={2023} }

@article{Darwish2024,
  title = {Blockchain for video watermarking: An enhanced copyright protection approach for video forensics based on perceptual hash function},
  volume = {19},
  ISSN = {1932-6203},
  url = {http://dx.doi.org/10.1371/journal.pone.0308451},
  DOI = {10.1371/journal.pone.0308451},
  number = {10},
  journal = {PLOS ONE},
  publisher = {Public Library of Science (PLoS)},
  author = {Darwish,  Saad Mohamed and Abu-Deif,  Mona Mahamod and Elkaffas,  Saleh Mesbah},
  editor = {Gupta,  Brij Bhooshan},
  year = {2024},
  month = oct,
  pages = {e0308451}
}

@article{Shao2024,
  title = {WFB: watermarking-based copyright protection framework for federated learning model via blockchain},
  volume = {14},
  ISSN = {2045-2322},
  url = {http://dx.doi.org/10.1038/s41598-024-70025-1},
  DOI = {10.1038/s41598-024-70025-1},
  number = {1},
  journal = {Scientific Reports},
  publisher = {Springer Science and Business Media LLC},
  author = {Shao,  Sujie and Wang,  Yue and Yang,  Chao and Liu,  Yan and Chen,  Xingyu and Qi,  Feng},
  year = {2024},
  month = aug 
}

@article{Lucchi2023,
  title = {ChatGPT: A Case Study on Copyright Challenges for Generative Artificial Intelligence Systems},
  ISSN = {2190-8249},
  url = {http://dx.doi.org/10.1017/err.2023.59},
  DOI = {10.1017/err.2023.59},
  journal = {European Journal of Risk Regulation},
  publisher = {Cambridge University Press (CUP)},
  author = {Lucchi,  Nicola},
  year = {2023},
  month = aug,
  pages = {1–23}
}

@article{kivus2024law,
  author = {Kivus, John T.},
  title = {Generative AI and Copyright Law: A Misalignment That Could Lead to the Privatization of Copyright Enforcement},
  journal = {North Carolina Journal of Law \& Technology},
  volume = {25},
  number = {3},
  month = {April},
  year = {2024},
  pages = {447--494}
}

@article{alhadeff2024fairuse,
  author = {Alhadeff, Jacob and Cuene, Cooper and Del Real, Max},
  title = {Limits of algorithmic fair use},
  journal = {Washington Journal of Law, Technology \& Arts},
  volume = {19},
  number = {1},
  pages = {1--53},
  year = {2024}
}

@inproceedings{liyanage2022benchmark,
      title={A Benchmark Corpus for the Detection of Automatically Generated Text in Academic Publications}, 
      author={Vijini Liyanage and Davide Buscaldi and Adeline Nazarenko},
      year={2022},
      pages = {4692-4700},
      booktitle = {13th Conference on Language Resources and Evaluation (LREC 2022)}
}

@misc {gptWikiIntro,
    author       = { {Bhat, Aaditya} },
    title        = { GPT-wiki-intro (Revision 0e458f5) },
    year         = 2023,
    url          = { https://huggingface.co/datasets/aadityaubhat/GPT-wiki-intro },
    doi          = { 10.57967/hf/0326 },
    publisher    = { Hugging Face }
}

@misc{coco,
  title={Common Objects in Context (COCO)},
  author={Lin, Tsung-Yi and Patterson, Genevieve and Ronchi, Matteo R. and Cui, Yin and Maire, Michael and Belongie, Serge and Bourdev, Lubomir and Girshick, Ross and Hays, James and Perona, Pietro and Ramanan, Deva and Zitnick, Larry and Dollár, Piotr},
  howpublished={\url{https://cocodataset.org/}},
  journal={cocodataset.org}
}

@article{microsoftCoco,
      title={Microsoft COCO: Common Objects in Context}, 
      author={Lin, Tsung-Yi and Maire, Michael and Belongie, Serge and Bourdev, Lubomir and Girshick, Ross and Hays, James and Perona, Pietro and Ramanan, Deva and Zitnick, C. Lawrence and Dollár, Piotr},
      year={2015},
      eprint={1405.0312},
      archivePrefix={arXiv},
      primaryClass={cs.CV}
}

@misc{ducru2024royalties,
      title={AI Royalties -- an IP Framework to Compensate Artists \& IP Holders for AI-Generated Content}, 
      author={Pablo Ducru and Jonathan Raiman and Ronaldo Lemos and Clay Garner and George He and Hanna Balcha and Gabriel Souto and Sergio Branco and Celina Bottino},
      year={2024},
      eprint={2406.11857},
      archivePrefix={arXiv},
      primaryClass={cs.CY},
      url={https://arxiv.org/abs/2406.11857}, 
}

@misc{wang2024econ,
      title={An Economic Solution to Copyright Challenges of Generative AI}, 
      author={Jiachen T. Wang and Zhun Deng and Hiroaki Chiba-Okabe and Boaz Barak and Weijie J. Su},
      year={2024},
      eprint={2404.13964},
      archivePrefix={arXiv},
      primaryClass={cs.LG},
      url={https://arxiv.org/abs/2404.13964}, 
}

@article{Geiger2024,
  title = {The forgotten creator: Towards a statutory remuneration right for machine learning of generative AI},
  volume = {52},
  ISSN = {0267-3649},
  url = {http://dx.doi.org/10.1016/j.clsr.2023.105925},
  DOI = {10.1016/j.clsr.2023.105925},
  journal = {Computer Law \&amp; Security Review},
  publisher = {Elsevier BV},
  author = {Geiger,  Christophe and Iaia,  Vincenzo},
  year = {2024},
  month = apr,
  pages = {105925}
}

@inproceedings{zhang2024lifecycle, 
   title={Privacy and Copyright Protection in Generative AI: A Lifecycle Perspective},
   volume={75},
   url={http://dx.doi.org/10.1145/3644815.3644952},
   DOI={10.1145/3644815.3644952},
   booktitle={Proceedings of the IEEE/ACM 3rd International Conference on AI Engineering - Software Engineering for AI},
   publisher={ACM},
   author={Zhang, Dawen and Xia, Boming and Liu, Yue and Xu, Xiwei and Hoang, Thong and Xing, Zhenchang and Staples, Mark and Lu, Qinghua and Zhu, Liming},
   year={2024},
}

@misc{lee2022deduplication,
      title={Deduplicating Training Data Makes Language Models Better}, 
      author={Katherine Lee and Daphne Ippolito and Andrew Nystrom and Chiyuan Zhang and Douglas Eck and Chris Callison-Burch and Nicholas Carlini},
      year={2022},
      eprint={2107.06499},
      archivePrefix={arXiv},
      primaryClass={cs.CL},
      url={https://arxiv.org/abs/2107.06499}, 
}

@article{RodriguezMaffioli2023,
  title = {Copyright in Generative AI training: Balancing Fair Use through Standardization and Transparency},
  ISSN = {1556-5068},
  url = {http://dx.doi.org/10.2139/ssrn.4579322},
  DOI = {10.2139/ssrn.4579322},
  journal = {SSRN Electronic Journal},
  publisher = {Elsevier BV},
  author = {Rodriguez Maffioli,  Daniel},
  year = {2023}
}

@article{zhang2024opt,
  title = {Opt-Out Implied Licenses in Copyright Law: From Search Engines to GPAI Models},
  volume = {73},
  ISSN = {2632-8550},
  url = {http://dx.doi.org/10.1093/grurint/ikae088},
  DOI = {10.1093/grurint/ikae088},
  number = {9},
  journal = {GRUR International},
  publisher = {Oxford University Press (OUP)},
  author = {Zhang,  Hongjiao and Li,  Yahong},
  year = {2024},
  month = jul,
  pages = {838–849}
}

@article{chan2024innov,
  title = {Balancing the Tradeoff between Regulation and Innovation for Artificial Intelligence: An Analysis of Top-down Command and Control and Bottom-up Self-Regulatory Approaches},
  ISSN = {0160-791X},
  url = {http://dx.doi.org/10.1016/j.techsoc.2024.102747},
  DOI = {10.1016/j.techsoc.2024.102747},
  journal = {Technology in Society},
  publisher = {Elsevier BV},
  author = {Chan,  Keith Jin Deng and Papyshev,  Gleb and Yarime,  Masaru},
  year = {2024},
  month = oct,
  pages = {102747}
}

@article{chesterman2024good,
  title={Good models borrow, great models steal: intellectual property rights and generative AI},
  author={Chesterman, Simon},
  journal={Policy and Society},
  pages={puae006},
  year={2024},
  publisher={Oxford University Press UK}
}

@misc{theMetOpen, title={The Met: Open Access}, url={https://www.metmuseum.org/about-the-met/policies-and-documents/open-access#get-started-header}, journal={The Metropolitan Museum of Art}, author={The Metropolitan Museum of Art} }

@article{flickr30k,
  author    = {Young, Peter and Lai, Alice and Hodosh, Micah and Hockenmaier, Julia},
  title     = {From image descriptions to visual denotations: New similarity metrics
               for semantic inference over event descriptions},
  journal   = {Transactions of the Association for Computational Linguistics},
  volume    = {2},
  pages     = {67--78},
  year      = {2014},
  url       = {https://www.aclweb.org/anthology/Q14-1006.pdf}
}

@article{openimages,
  author    = {Kuznetsova, Alina and Rom, Hassan and Alldrin, Neil and Uijlings, Jasper and Krasin, Ivan and Pont-Tuset, Jordi and Kamali, Shahab and Popov, Stefan and Malloci, Matteo and Duerig, Tom and Ferrari, Vittorio},
  title     = {{The Open Images Dataset V4:} Unified image classification, object
               detection, and visual relationship detection at scale},
  journal   = {CoRR},
  volume    = {abs/1811.00982},
  year      = {2018},
  url       = {http://arxiv.org/abs/1811.00982}
}

@misc{hfWikiart, url={https://huggingface.co/datasets/huggan/wikiart}, author={WikiArt}, journal={huggingface.co}, year={2022}, month={Dec} }

@misc{wikitextDataset, title={WikiText-103 Dataset}, url={https://huggingface.co/datasets/wikitext}, publisher={Hugging Face}, author={Merity, Stephen and Xiong, Caiming and Bradbury, James and Socher, Richard}, year={2016}, month={Sep} }

@misc{opensubtitleDataset, url={https://huggingface.co/datasets/open_subtitles}, author = {Lison, Pierre  and
      Tiedemann, Jörg}, publisher={Hugging Face}, year={2021}, month={May} }

@misc{bookcorpusDataset, author = {Bandy, John and Vincent, Nicholas}, url={https://huggingface.co/datasets/bookcorpus}, publisher={Hugging Face}, year={2021} }

@misc{jstorDataset, title={Text-mining support}, url={https://about.jstor.org/whats-in-jstor/text-mining-support/}, journal={JSTOR}, author={JSTOR} }

@inproceedings{langModelPlag, title={Do Language Models Plagiarize?}, DOI={https://doi.org/10.1145/3543507.3583199}, booktitle={WWW ’23: Proceedings of the ACM Web Conference}, author={Lee, Jooyoung and Le, Thai and Chen, Jinghui and Lee, Dongwon}, year={2023}, month={Apr}, pages={3637–3647} }

@article{deepCNN, title={ImageNet Classification with Deep Convolutional Neural Networks}, volume={60}, url={https://proceedings.neurips.cc/paper_files/paper/2012/file/c399862d3b9d6b76c8436e924a68c45b-Paper.pdf}, DOI={https://doi.org/10.1145/3065386}, number={6}, journal={Communications of the ACM}, author={Krizhevsky, Alex and Sutskever, Ilya and Hinton, Geoffrey E.}, year={2012}, month={May}, pages={84–90} }

@misc{fairUseDef,
  title = {Fair Use},
  author = {{Copyright Advisory Services}},
  publisher = {Columbia University Library},
  url = {https://copyright.columbia.edu/basics/fair-use.html},
  note = {Accessed: 2024-10-02}
}

@article{Opderbeck2024,
  title = {Copyright in AI Training Data: A Human-Centered Approach},
  ISSN = {1556-5068},
  url = {http://dx.doi.org/10.2139/ssrn.4679299},
  DOI = {10.2139/ssrn.4679299},
  journal = {SSRN Electronic Journal},
  publisher = {Elsevier BV},
  author = {Opderbeck,  David W.},
  year = {2024}
}

@misc{gozalobrizuela2023,
      title={ChatGPT is not all you need. A State of the Art Review of large Generative AI models}, 
      author={Roberto Gozalo-Brizuela and Eduardo C. Garrido-Merchan},
      year={2023},
      eprint={2301.04655},
      archivePrefix={arXiv},
      primaryClass={cs.LG},
      url={https://arxiv.org/abs/2301.04655}, 
}

@article{copybench,
      title={CopyBench: Measuring Literal and Non-Literal Reproduction of Copyright-Protected Text in Language Model Generation}, 
      author={Tong Chen and Akari Asai and Niloofar Mireshghallah and Sewon Min and James Grimmelmann and Yejin Choi and Hannaneh Hajishirzi and Luke Zettlemoyer and Pang Wei Koh},
      year={2024},
      eprint={2407.07087},
      archivePrefix={arXiv},
      primaryClass={cs.CL},
      url={https://arxiv.org/abs/2407.07087}, 
}

@misc{lee2024supplychain,
      title={Talkin' 'Bout AI Generation: Copyright and the Generative-AI Supply Chain}, 
      author={Katherine Lee and A. Feder Cooper and James Grimmelmann},
      year={2024},
      eprint={2309.08133},
      archivePrefix={arXiv},
      primaryClass={cs.CY},
      url={https://arxiv.org/abs/2309.08133}, 
}

@article{HernandezSuarez2023,
  title = {Methodological Approach for Identifying Websites with Infringing Content via Text Transformers and Dense Neural Networks},
  volume = {15},
  ISSN = {1999-5903},
  url = {http://dx.doi.org/10.3390/fi15120397},
  DOI = {10.3390/fi15120397},
  number = {12},
  journal = {Future Internet},
  publisher = {MDPI AG},
  author = {Hernandez-Suarez,  Aldo and Sanchez-Perez,  Gabriel and Toscano-Medina,  Linda Karina and Perez-Meana,  Hector Manuel and Portillo-Portillo,  Jose and Olivares-Mercado,  Jesus},
  year = {2023},
  month = dec,
  pages = {397}
}

@article{Xiao2012tchebi,
  title = {Radial Tchebichef moment invariants for image recognition},
  volume = {23},
  ISSN = {1047-3203},
  url = {http://dx.doi.org/10.1016/j.jvcir.2011.11.008},
  DOI = {10.1016/j.jvcir.2011.11.008},
  number = {2},
  journal = {Journal of Visual Communication and Image Representation},
  publisher = {Elsevier BV},
  author = {Xiao,  Bin and Ma,  Jian-Feng and Cui,  Jiang-Tao},
  year = {2012},
  month = feb,
  pages = {381–386}
}

@misc{xu2024license,
      title={A First Look at License Compliance Capability of LLMs in Code Generation}, 
      author={Weiwei Xu and Kai Gao and Hao He and Minghui Zhou},
      year={2024},
      eprint={2408.02487},
      archivePrefix={arXiv},
      primaryClass={cs.SE},
      url={https://arxiv.org/abs/2408.02487}, 
}

@misc{duarte2024decop,
      title={DE-COP: Detecting Copyrighted Content in Language Models Training Data}, 
      author={André V. Duarte and Xuandong Zhao and Arlindo L. Oliveira and Lei Li},
      year={2024},
      eprint={2402.09910},
      archivePrefix={arXiv},
      primaryClass={cs.CL},
      url={https://arxiv.org/abs/2402.09910}, 
}

@misc{zhou2023copyscope,
      title={CopyScope: Model-level Copyright Infringement Quantification in the Diffusion Workflow}, 
      author={Junlei Zhou and Jiashi Gao and Ziwei Wang and Xuetao Wei},
      year={2023},
      eprint={2311.12847},
      archivePrefix={arXiv},
      primaryClass={cs.CV},
      url={https://arxiv.org/abs/2311.12847}, 
}

@misc{zhao2024advwatermark,
      title={Invisible Image Watermarks Are Provably Removable Using Generative AI}, 
      author={Xuandong Zhao and Kexun Zhang and Zihao Su and Saastha Vasan and Ilya Grishchenko and Christopher Kruegel and Giovanni Vigna and Yu-Xiang Wang and Lei Li},
      year={2024},
      eprint={2306.01953},
      archivePrefix={arXiv},
      primaryClass={cs.CR},
      url={https://arxiv.org/abs/2306.01953}, 
}

@inproceedings{yew2024liability,
author = {Yew, Rui-Jie},
title = {Break It 'Til You Make It: An Exploration of the Ramifications of Copyright Liability Under a Pre-training Paradigm of AI Development},
year = {2024},
isbn = {9798400703331},
publisher = {Association for Computing Machinery},
address = {New York, NY, USA},
url = {https://doi.org/10.1145/3614407.3643707},
doi = {10.1145/3614407.3643707},
booktitle = {Proceedings of the Symposium on Computer Science and Law},
pages = {64–72},
numpages = {9},
location = {Boston, MA, USA},
series = {CSLAW '24}
}

@misc{kim2024jail,
      title={Automatic Jailbreaking of the Text-to-Image Generative AI Systems}, 
      author={Minseon Kim and Hyomin Lee and Boqing Gong and Huishuai Zhang and Sung Ju Hwang},
      year={2024},
      eprint={2405.16567},
      archivePrefix={arXiv},
      primaryClass={cs.AI},
      url={https://arxiv.org/abs/2405.16567}, 
}

@misc{deng2024defense,
      title={A Survey of Defenses against AI-generated Visual Media: Detection, Disruption, and Authentication}, 
      author={Jingyi Deng and Chenhao Lin and Zhengyu Zhao and Shuai Liu and Qian Wang and Chao Shen},
      year={2024},
      eprint={2407.10575},
      archivePrefix={arXiv},
      primaryClass={cs.CV},
      url={https://arxiv.org/abs/2407.10575}, 
}

@misc{sai2024blockchaincontracts,
      title={Is Your AI Truly Yours? Leveraging Blockchain for Copyrights, Provenance, and Lineage}, 
      author={Yilin Sai and Qin Wang and Guangsheng Yu and H. M. N. Dilum Bandara and Shiping Chen},
      year={2024},
      eprint={2404.06077},
      archivePrefix={arXiv},
      primaryClass={cs.CR},
      url={https://arxiv.org/abs/2404.06077}, 
}

@article{Zhou_2022_dedup,
   title={Serving deep learning models with deduplication from relational databases},
   volume={15},
   ISSN={2150-8097},
   url={http://dx.doi.org/10.14778/3547305.3547325},
   DOI={10.14778/3547305.3547325},
   number={10},
   journal={Proceedings of the VLDB Endowment},
   publisher={Association for Computing Machinery (ACM)},
   author={Zhou, Lixi and Chen, Jiaqing and Das, Amitabh and Min, Hong and Yu, Lei and Zhao, Ming and Zou, Jia},
   year={2022},
   month=jun, pages={2230–2243} }

@article{RasinaBegum2021,
  title = {SEEDDUP: A Three-Tier SEcurE Data DedUPlication Architecture-Based Storage and Retrieval for Cross-Domains Over Cloud},
  volume = {69},
  ISSN = {0974-780X},
  url = {http://dx.doi.org/10.1080/03772063.2021.1886882},
  DOI = {10.1080/03772063.2021.1886882},
  number = {4},
  journal = {IETE Journal of Research},
  publisher = {Informa UK Limited},
  author = {Rasina Begum,  B. and Chitra,  P.},
  year = {2021},
  month = feb,
  pages = {2224–2241}
}

@misc{li2024gendedup,
      title={Generative Deduplication For Socia Media Data Selection}, 
      author={Xianming Li and Jing Li},
      year={2024},
      eprint={2401.05883},
      archivePrefix={arXiv},
      primaryClass={cs.CL},
      url={https://arxiv.org/abs/2401.05883}, 
}

@article{Cai2023,
  title = {Image neural style transfer: A review},
  volume = {108},
  ISSN = {0045-7906},
  url = {http://dx.doi.org/10.1016/j.compeleceng.2023.108723},
  DOI = {10.1016/j.compeleceng.2023.108723},
  journal = {Computers and Electrical Engineering},
  publisher = {Elsevier BV},
  author = {Cai,  Qiang and Ma,  Mengxu and Wang,  Chen and Li,  Haisheng},
  year = {2023},
  month = may,
  pages = {108723}
}

@inproceedings{Li2024styletrans,
  title = {Neural Style Protection: Counteracting Unauthorized Neural Style Transfer},
  volume = {34},
  url = {http://dx.doi.org/10.1109/WACV57701.2024.00392},
  DOI = {10.1109/wacv57701.2024.00392},
  booktitle = {2024 IEEE/CVF Winter Conference on Applications of Computer Vision (WACV)},
  publisher = {IEEE},
  author = {Li,  Yaxin and Ren,  Jie and Xu,  Han and Liu,  Hui},
  year = {2024},
  month = jan,
  pages = {3954–3963}
}

@article{Rosati2024,
  title = {Infringing AI: Liability for AI-generated outputs under international,  EU,  and UK copyright law},
  url = {http://dx.doi.org/10.2139/ssrn.4946312},
  DOI = {10.2139/ssrn.4946312},
  publisher = {Elsevier BV},
  author = {Rosati,  Eleonora},
  year = {2024}
}

@article{Felzmann2020,
  title = {Towards Transparency by Design for Artificial Intelligence},
  volume = {26},
  ISSN = {1471-5546},
  url = {http://dx.doi.org/10.1007/s11948-020-00276-4},
  DOI = {10.1007/s11948-020-00276-4},
  number = {6},
  journal = {Science and Engineering Ethics},
  publisher = {Springer Science and Business Media LLC},
  author = {Felzmann,  Heike and Fosch-Villaronga,  Eduard and Lutz,  Christoph and Tamò-Larrieux,  Aurelia},
  year = {2020},
  month = nov,
  pages = {3333–3361}
}

@misc{chmielinski2022nutrition,
      title={The Dataset Nutrition Label (2nd Gen): Leveraging Context to Mitigate Harms in Artificial Intelligence}, 
      author={Kasia S. Chmielinski and Sarah Newman and Matt Taylor and Josh Joseph and Kemi Thomas and Jessica Yurkofsky and Yue Chelsea Qiu},
      year={2022},
      eprint={2201.03954},
      archivePrefix={arXiv},
      primaryClass={cs.LG},
      url={https://arxiv.org/abs/2201.03954}, 
}

@misc{si2024nutrition,
      title={A Solution toward Transparent and Practical AI Regulation: Privacy Nutrition Labels for Open-source Generative AI-based Applications}, 
      author={Meixue Si and Shidong Pan and Dianshu Liao and Xiaoyu Sun and Zhen Tao and Wenchang Shi and Zhenchang Xing},
      year={2024},
      eprint={2407.15407},
      archivePrefix={arXiv},
      primaryClass={cs.CR},
      url={https://arxiv.org/abs/2407.15407}, 
}

@inproceedings{Pushkarna2022,
  series = {FAccT ’22},
  title = {Data Cards: Purposeful and Transparent Dataset Documentation for Responsible AI},
  url = {http://dx.doi.org/10.1145/3531146.3533231},
  DOI = {10.1145/3531146.3533231},
  booktitle = {2022 ACM Conference on Fairness,  Accountability,  and Transparency},
  publisher = {ACM},
  author = {Pushkarna,  Mahima and Zaldivar,  Andrew and Kjartansson,  Oddur},
  year = {2022},
  month = jun,
  collection = {FAccT ’22}
}

@misc{manheim2024audit,
      title={The Necessity of AI Audit Standards Boards}, 
      author={David Manheim and Sammy Martin and Mark Bailey and Mikhail Samin and Ross Greutzmacher},
      year={2024},
      eprint={2404.13060},
      archivePrefix={arXiv},
      primaryClass={cs.CY},
      url={https://arxiv.org/abs/2404.13060}, 
}

@misc{birhane2024audit,
      title={AI auditing: The Broken Bus on the Road to AI Accountability}, 
      author={Abeba Birhane and Ryan Steed and Victor Ojewale and Briana Vecchione and Inioluwa Deborah Raji},
      year={2024},
      eprint={2401.14462},
      archivePrefix={arXiv},
      primaryClass={cs.CY},
      url={https://arxiv.org/abs/2401.14462}, 
}

@inproceedings{Casper2024,
  series = {FAccT ’24},
  title = {Black-Box Access is Insufficient for Rigorous AI Audits},
  url = {http://dx.doi.org/10.1145/3630106.3659037},
  DOI = {10.1145/3630106.3659037},
  booktitle = {The 2024 ACM Conference on Fairness,  Accountability,  and Transparency},
  publisher = {ACM},
  author = {Casper,  Stephen and Ezell,  Carson and Siegmann,  Charlotte and Kolt,  Noam and Curtis,  Taylor Lynn and Bucknall,  Benjamin and Haupt,  Andreas and Wei,  Kevin and Scheurer,  Jérémy and Hobbhahn,  Marius and Sharkey,  Lee and Krishna,  Satyapriya and Von Hagen,  Marvin and Alberti,  Silas and Chan,  Alan and Sun,  Qinyi and Gerovitch,  Michael and Bau,  David and Tegmark,  Max and Krueger,  David and Hadfield-Menell,  Dylan},
  year = {2024},
  month = jun,
  pages = {2254–2272},
  collection = {FAccT ’24}
}

@misc{xie2024mugc,
      title={MUGC: Machine Generated versus User Generated Content Detection}, 
      author={Yaqi Xie and Anjali Rawal and Yujing Cen and Dixuan Zhao and Sunil K Narang and Shanu Sushmita},
      year={2024},
      eprint={2403.19725},
      archivePrefix={arXiv},
      primaryClass={cs.CL},
      url={https://arxiv.org/abs/2403.19725}, 
}

@article{Schneider2024,
  title = {Explainable Generative AI (GenXAI): a survey,  conceptualization,  and research agenda},
  volume = {57},
  ISSN = {1573-7462},
  url = {http://dx.doi.org/10.1007/s10462-024-10916-x},
  DOI = {10.1007/s10462-024-10916-x},
  number = {11},
  journal = {Artificial Intelligence Review},
  publisher = {Springer Science and Business Media LLC},
  author = {Schneider,  Johannes},
  year = {2024},
  month = sep 
}

@article{Ohm2024,
  title = {Focusing on Fine-Tuning: Understanding the Four Pathways for Shaping Generative AI},
  ISSN = {1556-5068},
  url = {http://dx.doi.org/10.2139/ssrn.4738261},
  DOI = {10.2139/ssrn.4738261},
  journal = {SSRN Electronic Journal},
  publisher = {Elsevier BV},
  author = {Ohm,  Paul},
  year = {2024}
}

@inproceedings{Zhong_2023, series={WWW ’23},
   title={Copyright Protection and Accountability of Generative AI: Attack, Watermarking and Attribution},
   url={http://dx.doi.org/10.1145/3543873.3587321},
   DOI={10.1145/3543873.3587321},
   booktitle={Companion Proceedings of the ACM Web Conference 2023},
   publisher={ACM},
   author={Zhong, Haonan and Chang, Jiamin and Yang, Ziyue and Wu, Tingmin and Mahawaga Arachchige, Pathum Chamikara and Pathmabandu, Chehara and Xue, Minhui},
   year={2023},
   month=apr, collection={WWW ’23} }

@misc{chu2023softmax,
      title={How to Protect Copyright Data in Optimization of Large Language Models?}, 
      author={Timothy Chu and Zhao Song and Chiwun Yang},
      year={2023},
      eprint={2308.12247},
      archivePrefix={arXiv},
      primaryClass={cs.LG},
      url={https://arxiv.org/abs/2308.12247}, 
}

@article{chang2024forgotten,
    author       = {Chang, Cheng-chi},
    title        = {When AI Remembers Too Much: Reinventing the Right to Be Forgotten for the Generative Age},
    journal      = {Washington Journal of Law, Technology \& Arts},
    volume       = {19},
    year         = {2024},
    url          = {https://digitalcommons.law.uw.edu/wjlta/vol19/iss3/2}
}

@article{henderson2023Foundation,
  author  = {Peter Henderson and Xuechen Li and Dan Jurafsky and Tatsunori Hashimoto and Mark A. Lemley and Percy Liang},
  title   = {Foundation Models and Fair Use},
  journal = {Journal of Machine Learning Research},
  year    = {2023},
  volume  = {24},
  number  = {400},
  pages   = {1--79},
  url     = {http://jmlr.org/papers/v24/23-0569.html}
}

@article{Preetha2023,
  title = {A Wavelet Optimized Video Copy Detection Using Content Fingerprinting},
  volume = {95},
  ISSN = {1939-8115},
  url = {http://dx.doi.org/10.1007/s11265-022-01830-y},
  DOI = {10.1007/s11265-022-01830-y},
  number = {2–3},
  journal = {Journal of Signal Processing Systems},
  publisher = {Springer Science and Business Media LLC},
  author = {Preetha,  S. and Bindu,  V. R.},
  year = {2023},
  month = feb,
  pages = {363–377}
}

@misc{shumailov2024ununlearning,
      title={UnUnlearning: Unlearning is not sufficient for content regulation in advanced generative AI}, 
      author={Ilia Shumailov and Jamie Hayes and Eleni Triantafillou and Guillermo Ortiz-Jimenez and Nicolas Papernot and Matthew Jagielski and Itay Yona and Heidi Howard and Eugene Bagdasaryan},
      year={2024},
      eprint={2407.00106},
      archivePrefix={arXiv},
      primaryClass={cs.LG},
      url={https://arxiv.org/abs/2407.00106}, 
}

@inproceedings{ning2021reg,
  author={Ning, Bowen and Niu, Baoning and Guan, Hu and Huang, Ying and Zhang, Shuwu},
  booktitle={2021 International Conference on Culture-oriented Science \& Technology (ICCST)}, 
  title={Research and Development of Copyright Registration and Monitoring System Based on Digital Watermarking and Fingerprint Technology}, 
  year={2021},
  volume={},
  number={},
  pages={354-358},
  doi={10.1109/ICCST53801.2021.00080}}

@misc{chen2024fingerprint,
      title={Digital Fingerprinting on Multimedia: A Survey}, 
      author={Wendi Chen and Wensheng Gan and Philip S. Yu},
      year={2024},
      eprint={2408.14155},
      archivePrefix={arXiv},
      primaryClass={cs.MM},
      url={https://arxiv.org/abs/2408.14155}, 
}

@mastersthesis{della_giustina_2024,
    author       = {Coelho Della Giustina, Amanda},
    title        = {Fair Compensation for Copyrighted Data Used in AI Training},
    school       = {Tilburg University},
    year         = {2024},
    type         = {Master's Thesis},
    note         = {LLM. Law \& Technology, supervised by Pratham Ajmera and Dr. Anuj Puri},
    institution  = {Tilburg Institute for Law, Technology, and Society (TILT)}
}

@inproceedings{balan2023opt,
author = {Balan, Kar and Gilbert, Andrew and Black, Alexander and Jenni, Simon and Parsons, Andy and Collomosse, John},
title = {DECORAIT - DECentralized Opt-in/out Registry for AI Training},
year = {2023},
isbn = {9798400704260},
publisher = {Association for Computing Machinery},
address = {New York, NY, USA},
url = {https://doi.org/10.1145/3626495.3626506},
doi = {10.1145/3626495.3626506},
booktitle = {Proceedings of the 20th ACM SIGGRAPH European Conference on Visual Media Production},
articleno = {4},
numpages = {10},
location = {London, United Kingdom},
series = {CVMP '23}
}

@techreport{keller2023opt,
    author    = {Keller, Paul and Warso, Zuzanna},
    title     = {Defining Best Practices for Opting Out of ML Training},
    institution = {Open Future},
    year      = {2023},
    month     = {September},
    day       = {28},
    type      = {Policy Brief},
    url       = {https://openfuture.eu/reports/defining-best-practices-for-opting-out-of-ml-training/}
}

@article{Pasquale2024,
  title = {Consent and Compensation: Resolving Generative AI’s Copyright Crisis},
  ISSN = {1556-5068},
  url = {http://dx.doi.org/10.2139/ssrn.4826695},
  DOI = {10.2139/ssrn.4826695},
  journal = {SSRN Electronic Journal},
  publisher = {Elsevier BV},
  author = {Pasquale,  Frank A. and Sun,  Haochen},
  year = {2024}
}

@article{ziaja2024opt,
  title = {The text and data mining opt-out in Article 4(3) CDSMD: Adequate veto right for rightholders or a suffocating blanket for European artificial intelligence innovations?},
  volume = {19},
  ISSN = {1747-1540},
  url = {http://dx.doi.org/10.1093/jiplp/jpae025},
  DOI = {10.1093/jiplp/jpae025},
  number = {5},
  journal = {Journal of Intellectual Property Law \& Practice},
  publisher = {Oxford University Press (OUP)},
  author = {Ziaja,  Gina Maria},
  year = {2024},
  month = feb,
  pages = {453–459}
}

@article{rafatijo2020software,
    author       = {Rafatijo, Homayoon and Crouch, Dennis},
    title        = {When Is Software Code Copyrightable? Is Its Unauthorized Copying Excusable as a Fair Use?},
    journal      = {Preview of United States Supreme Court Cases},
    volume       = {47},
    number       = {6},
    pages        = {14},
    year         = {2020}
}

@article{kang2023multimodal,
  title = {Beyond ChatGPT: Multimodal generative AI for L2 writers},
  volume = {62},
  ISSN = {1060-3743},
  url = {http://dx.doi.org/10.1016/j.jslw.2023.101070},
  DOI = {10.1016/j.jslw.2023.101070},
  journal = {Journal of Second Language Writing},
  publisher = {Elsevier BV},
  author = {Kang,  Joohoon and Yi,  Youngjoo},
  year = {2023},
  month = dec,
  pages = {101070}
}

@article{lu2024multimodal,
  title = {A multimodal generative AI copilot for human pathology},
  volume = {634},
  ISSN = {1476-4687},
  url = {http://dx.doi.org/10.1038/s41586-024-07618-3},
  DOI = {10.1038/s41586-024-07618-3},
  number = {8033},
  journal = {Nature},
  publisher = {Springer Science and Business Media LLC},
  author = {Lu,  Ming Y. and Chen,  Bowen and Williamson,  Drew F. K. and Chen,  Richard J. and Zhao,  Melissa and Chow,  Aaron K. and Ikemura,  Kenji and Kim,  Ahrong and Pouli,  Dimitra and Patel,  Ankush and Soliman,  Amr and Chen,  Chengkuan and Ding,  Tong and Wang,  Judy J. and Gerber,  Georg and Liang,  Ivy and Le,  Long Phi and Parwani,  Anil V. and Weishaupt,  Luca L. and Mahmood,  Faisal},
  year = {2024},
  month = jun,
  pages = {466–473}
}

@article{liang2024multimodal,
author = {Liang, Paul Pu and Zadeh, Amir and Morency, Louis-Philippe},
title = {Foundations \& Trends in Multimodal Machine Learning: Principles, Challenges, and Open Questions},
year = {2024},
issue_date = {October 2024},
publisher = {Association for Computing Machinery},
address = {New York, NY, USA},
volume = {56},
number = {10},
issn = {0360-0300},
url = {https://doi.org/10.1145/3656580},
doi = {10.1145/3656580},
journal = {ACM Comput. Surv.},
month = jun,
articleno = {264},
numpages = {42}
}

@misc{cao2023survey,
      title={A Comprehensive Survey of AI-Generated Content (AIGC): A History of Generative AI from GAN to ChatGPT}, 
      author={Yihan Cao and Siyu Li and Yixin Liu and Zhiling Yan and Yutong Dai and Philip S. Yu and Lichao Sun},
      year={2023},
      eprint={2303.04226},
      archivePrefix={arXiv},
      primaryClass={cs.AI},
      url={https://arxiv.org/abs/2303.04226}, 
}

@misc{xu2023small,
      title={Small Models are Valuable Plug-ins for Large Language Models}, 
      author={Canwen Xu and Yichong Xu and Shuohang Wang and Yang Liu and Chenguang Zhu and Julian McAuley},
      year={2023},
      eprint={2305.08848},
      archivePrefix={arXiv},
      primaryClass={cs.CL},
      url={https://arxiv.org/abs/2305.08848}, 
}

@misc{hsieh2023small,
      title={Distilling Step-by-Step! Outperforming Larger Language Models with Less Training Data and Smaller Model Sizes}, 
      author={Cheng-Yu Hsieh and Chun-Liang Li and Chih-Kuan Yeh and Hootan Nakhost and Yasuhisa Fujii and Alexander Ratner and Ranjay Krishna and Chen-Yu Lee and Tomas Pfister},
      year={2023},
      eprint={2305.02301},
      archivePrefix={arXiv},
      primaryClass={cs.CL},
      url={https://arxiv.org/abs/2305.02301}, 
}

@misc{li2024nsmall,
      title={Symbolic Chain-of-Thought Distillation: Small Models Can Also "Think" Step-by-Step}, 
      author={Liunian Harold Li and Jack Hessel and Youngjae Yu and Xiang Ren and Kai-Wei Chang and Yejin Choi},
      year={2024},
      eprint={2306.14050},
      archivePrefix={arXiv},
      primaryClass={cs.CL},
      url={https://arxiv.org/abs/2306.14050}, 
}

@misc{cooper2024mem,
      title={The Files are in the Computer: Copyright, Memorization, and Generative AI}, 
      author={A. Feder Cooper and James Grimmelmann},
      year={2024},
      eprint={2404.12590},
      archivePrefix={arXiv},
      primaryClass={cs.CY},
      url={https://arxiv.org/abs/2404.12590}, 
}

@misc{ross2024mem,
      title={A Geometric Framework for Understanding Memorization in Generative Models}, 
      author={Brendan Leigh Ross and Hamidreza Kamkari and Tongzi Wu and Rasa Hosseinzadeh and Zhaoyan Liu and George Stein and Jesse C. Cresswell and Gabriel Loaiza-Ganem},
      year={2024},
      eprint={2411.00113},
      archivePrefix={arXiv},
      primaryClass={stat.ML},
      url={https://arxiv.org/abs/2411.00113}, 
}

@misc{zhang2022pal,
      title={Perceptual Artifacts Localization for Inpainting}, 
      author={Lingzhi Zhang and Yuqian Zhou and Connelly Barnes and Sohrab Amirghodsi and Zhe Lin and Eli Shechtman and Jianbo Shi},
      year={2022},
      eprint={2208.03357},
      archivePrefix={arXiv},
      primaryClass={cs.CV},
      url={https://arxiv.org/abs/2208.03357}, 
}

@misc{zhang2023pal,
      title={Perceptual Artifacts Localization for Image Synthesis Tasks}, 
      author={Lingzhi Zhang and Zhengjie Xu and Connelly Barnes and Yuqian Zhou and Qing Liu and He Zhang and Sohrab Amirghodsi and Zhe Lin and Eli Shechtman and Jianbo Shi},
      year={2023},
      eprint={2310.05590},
      archivePrefix={arXiv},
      primaryClass={cs.CV},
      url={https://arxiv.org/abs/2310.05590}, 
}

@misc{aboutalebi2024,
      title={DeepfakeArt Challenge: A Benchmark Dataset for Generative AI Art Forgery and Data Poisoning Detection}, 
      author={Hossein Aboutalebi and Dayou Mao and Rongqi Fan and Carol Xu and Chris He and Alexander Wong},
      year={2024},
      eprint={2306.01272},
      archivePrefix={arXiv},
      primaryClass={cs.CV},
      url={https://arxiv.org/abs/2306.01272}, 
}

@inproceedings{shen2022card,
author = {Shen, Hong and Wang, Leijie and Deng, Wesley H. and Brusse, Ciell and Velgersdijk, Ronald and Zhu, Haiyi},
title = {The Model Card Authoring Toolkit: Toward Community-centered, Deliberation-driven AI Design},
year = {2022},
isbn = {9781450393522},
publisher = {Association for Computing Machinery},
address = {New York, NY, USA},
url = {https://doi.org/10.1145/3531146.3533110},
doi = {10.1145/3531146.3533110},
booktitle = {Proceedings of the 2022 ACM Conference on Fairness, Accountability, and Transparency},
pages = {440–451},
numpages = {12},
location = {Seoul, Republic of Korea},
series = {FAccT '22}
}

@misc{icoTookit,
    title = {{AI and Data Protection Risk Toolkit}},
    author = {{UK Information Commissioner's Office}},
    year = {2024},
    month = may,
    day = {24},
    url = {https://ico.org.uk/for-organisations/uk-gdpr-guidance-and-resources/artificial-intelligence/guidance-on-ai-and-data-protection/ai-and-data-protection-risk-toolkit/},
    note = {Accessed: 2024-11-05}
}

@misc{ethicsAlgToolkit,
    title = {{Ethics \& Algorithms Toolkit}},
    author = {David Anderson and Joy Bonaguro and Miriam McKinney and Andrew Nicklin and Jane Wiseman},
    organization = {{Center for Government Excellence at Johns Hopkins University}},
    year = {2018},
    url = {https://ethicstoolkit.ai/},
    note = {Accessed: 2024-11-05}
}

\end{document}